\documentclass[sn-mathphys,Numbered]{sn-jnl}% Math and Physical Sciences Reference Style
\usepackage{graphicx}%
\usepackage{multirow}%
\usepackage{amsmath,amssymb,amsfonts}%
\usepackage{amsthm}%
\usepackage{mathrsfs}%
\usepackage[title]{appendix}%
\usepackage{xcolor}%
\usepackage{textcomp}%
\usepackage{manyfoot}%
\usepackage{booktabs}%
\usepackage{algorithm}%
\usepackage{algorithmicx}%
\usepackage{algpseudocode}%
\usepackage{listings}%

\theoremstyle{thmstyleone}%
\theoremstyle{thmstyletwo}%

\theoremstyle{thmstylethree}%

\begin{document}

\title[Research and application of Transformer based anomaly detection model: A literature review]{Research and application of Transformer based anomaly detection model: A literature review}


\author[1]{\fnm{Mingrui} \sur{Ma}}\email{m202271767@hust.edu.cn}

\author*[1]{\fnm{Lansheng} \sur{Han}}\email{1998010309@hust.edu.cn}

\author[2]{\fnm{Chunjie} \sur{Zhou}}\email{cjiezhou@hust.edu.cn}

\affil[1]{\orgdiv{Hubei Key Laboratory of Distributed System Security, Hubei
Engineering Research Center on Big Data Security, School of Cyber Science and
Engineering}, \orgname{Huazhong University of Science and Technology}, \orgaddress{\city{Wuhan}, \postcode{430074}, \state{Hubei}, \country{China}}}

\affil[2]{\orgdiv{The Key Laboratory of Ministry of Education for Image Processing and Intelligent Control, School of Artificial Intelligence and Automation}, \orgname{Huazhong University of Science and Technology}, \orgaddress{\city{Wuhan}, \postcode{430074}, \state{Hubei}, \country{China}}}


\abstract{Transformer, as one of the most advanced neural network models in Natural Language Processing (NLP), exhibits diverse applications in the field of anomaly detection. To inspire research on Transformer-based anomaly detection, this review offers a fresh perspective on the concept of anomaly detection. We explore the current challenges of anomaly detection and provide detailed insights into the operating principles of Transformer and its variants in anomaly detection tasks. Additionally, we delineate various application scenarios for Transformer-based anomaly detection models and discuss the datasets and evaluation metrics employed. Furthermore, this review highlights the key challenges in Transformer-based anomaly detection research and conducts a comprehensive analysis of future research trends in this domain. The review includes an extensive compilation of over 100 core references related to Transformer-based anomaly detection. To the best of our knowledge, this is the first comprehensive review that focuses on the research related to Transformer in the context of anomaly detection. We hope that this paper can provide detailed technical information to researchers interested in Transformer-based anomaly detection tasks.}




\maketitle

\section{Introduction}\label{sec1}

Anomaly detection \cite{RF1} is the task of identifying anomalous data that are statistically different from normal instances in terms of numerical characteristics. Anomaly detection has a wide range of applications, including but not limited to system logs \cite{RF2}, industrial control systems (ICS) \cite{RF3}, autonomous driving systems \cite{RF4}, medical diagnostics \cite{RF5}, spacecraft \cite{RF6}, etc. There are many factors leading to system failures, such as system errors, human factors, natural disasters, malicious intrusions, etc. The purpose of anomaly detection is to precisely identify system failures and guarantee the stable operation of the system. 

According to the timeline, the technical development of anomaly detection has gone through statistical methods, machine learning-based algorithms, neural networks, and deep learning-based methods. Experimental results show that the methods based on neural networks and deep learning can achieve better detection performance than statistical methods and machine learning algorithms due to their outstanding capabilities in learning expressive representations of complex data. (i.e., high-dimensional data, spatial data, etc.) At present, a variety of neural network architectures have been applied to the task of anomaly detection, such as Convolutional Neural Network (CNN) \cite{RF7}, Recurrent Neural Network (RNN) \cite{RF8}, Encoder-Decoder structure \cite{RF9}, Transformer \cite{RF2} and Generative Adversarial Network (GAN) \cite{RF10}. 

With the diversification of anomaly detection application domains, variants of mainstream neural networks have been proposed to achieve better performance in domain-specific anomaly detection tasks. However, current datasets tend to have quantities of positive samples (normal samples) and a very small number of negative samples (anomalous samples), resulting in serious data imbalance \cite{RF11}. This also leads to the fact that most methods based on neural networks and deep learning adopt unsupervised learning approaches for model training. Therefore, before Transformer was applied to the anomaly detection task, the industry generally considered Variational Autoencoder (VAE) \cite{RF12} and GAN, which are capable of distribution fitting, were the best methods for anomaly detection. VAE is used to minimize the log-likelihood lower bound on the data, while GAN is to achieve a balance between the generator and the discriminator. Although the GAN network \cite{RF10} has shown superior performance in the task of anomaly detection, it also suffers from training instability. (i.e., difficulty in entering Nash equilibrium) To this end, many variants of GAN-based neural networks have also been proposed, such as WGAN \cite{RF13}. 

With Transformer \cite{RF14} being proposed in 2017, its unique attention mechanism with excellent contextual feature extraction performance makes it not only competent for NLP tasks but also for a wide range of transfer learning tasks. Google released BERT \cite{RF15} in 2019, an important variant of Transformer. BERT introduces Bi-directional operations based on Vanilla Transformer and uses the pre-training mechanism to further enhance model performance and reduce training costs. BERT model has also achieved state-of-the-art (SOTA) results in 11 NLP tasks such as Named Entity Recognition (NER), Machine Translation (MT), etc. Transformer essentially changes the linear operation structure of the previous RNNs (i.e. LSTM, GRU, etc.), and realizes parallel operation through pure attention mechanism, which greatly improves the operation efficiency and has further contextual memory ability. 

Therefore, Transformer model is well suited for anomaly detection tasks with strong contextual information associations such as serialized data. In related studies, researchers have employed various approaches for anomaly detection. Some utilize the Vanilla Transformer model, while others directly adopt the BERT model \cite{RF16}. Additionally, some researchers modify the Transformer structure, such as using Vision Transformer (ViT) \cite{RF17}, and others integrate Transformer with auxiliary methods, as seen in TranAD \cite{RF18}. TranAD integrates the generative adversarial training approach from GAN into the Transformer model through refactoring and applying it to the multivariate time series data (MTS) anomaly detection tasks. This idea has also been adopted by Adformer\cite{ZENG2023244}.

In this comprehensive review, we will provide an in-depth analysis of the theoretical advancements and application progress of Transformers in anomaly detection tasks over the past seven years. Furthermore, we will identify the existing open issues, research gaps, and future trends in Transformer-based anomaly detection tasks. To the best of our knowledge, this will be the first work to comprehensively summarize the research progress of Transformers in the field of anomaly detection, offering valuable insights for future investigations. This helps the reader to gain key insights into relevant research and provides illuminating thoughts and genuine open opportunities on some of the difficulties and challenges in Transformer-based anomaly detection. 

The rest of the paper is organized as follows: Section \ref{2} provides a comprehensive overview and analysis on the existing definitions of anomaly detection concepts from various research sources. Section \ref{3} delves into the interrelationships between different learning paradigms, offering a consolidated summary. Section \ref{4} presents a systematic categorization and summary of the operational principles of Transformer-related models, along with their application perspectives in the realm of anomaly detection. Section \ref{5} conducts an examination and classification of relevant research based on different datatypes. Section \ref{6} critically evaluates the evaluation indexes and datasets employed in current studies, highlighting the strengths and weaknesses of existing index systems and widely adopted open-access datasets across diverse application scenarios. Section \ref{7} outlines the overarching challenges of anomaly detection, specifically emphasizing the core issues that warrant consideration when utilizing Transformers, while also discussing future research trends within this domain. Section \ref{8} provides a concise summary of our work.

\section{Concepts of anomaly detection}\label{2}

Anomaly detection is a relatively broad concept. Many studies have proposed very similar concepts, such as outliers \cite{RF19}, out of distributions (OOD) \cite{RF20}, bias \cite{RF21}, and novelty \cite{RF22}. Although the concepts are different, we can divide them into 3 categories as a whole: anomaly, outliers/OOD, and novelty. In general, anomalies are considered objects (behaviors, patterns, distributions, cases, etc.) that are inconsistent with the data in the normal range of human cognition (normal instances, normal distributions, normal states, characteristics of most objects, etc.) \cite{RF1, RF23, RF24}. Anomalies can be biased from the majority of observations, behaviors that do not conform to a clear definition, etc. Currently, many scholars have tried to give different classifications for anomaly detection, but the corresponding classification methods vary due to the actual needs and application scenarios. For example, Ma et al. \cite{RF25} believed that social spammers, misinformation, fraudsters, network intruders, etc. all fall into the category of anomalies. Bulusu et al. \cite{RF26} further divided anomalies into intentional anomalies and unintentional anomalies. However, this division is based on the following conditions:

\begin{enumerate}
    \item Normal samples have the same feature distribution in the latent space;
    \item The distribution of anomalous samples is very different from that of normal samples.
\end{enumerate}

In fact, different scientific scholars have contradictory definitions of anomalies. Boukerche et al. \cite{RF27} first identified and proposed the subtle differences in the concepts of outliers and anomalies, but they still used the two concepts confusingly in their paper. In addition, different researchers disagree greatly about whether novelty belongs to the category of anomalies.
Miljkovic et al. \cite{RF28} argued that novelty belongs to the category of anomaly because novelty belongs to the data of the corresponding region in the feature space. While Chandola et al. \cite{RF1} suggested that novelty does not belong to anomalies. The fact remains, however, that there is a large number of research methods used for novelty detection that can also be applied to the task of anomaly detection. Ander et al. \cite{RF29} emphasized the differences between related concepts, including anomalies, novelty, outlier, and rare event detection terms. They indicated that these related concepts can be applied to the task of supervised learning. However, they did not further explain whether these related concepts can be extended to other application scenarios such as unsupervised learning. In Table \ref{Table1}, we summarize the definitions of anomalies in different literature. In summary, there is no general, formalized, and rigorous definition of anomaly in the field of anomaly detection, and researchers have not reached a consensus on the concept of anomaly detection.
 
\begin{table}[h]\footnotesize
    \caption{Definitions of anomalies in different research results}
    \begin{tabular*}{\textwidth}{cp{4cm}p{6.6cm}}
    \toprule
    Concepts & Title & Definitions \\
    \midrule
    Anomaly & Anomaly Detection: A Survey \cite{RF1} & Anomalies are patterns in data that do not conform to a well-defined notion of normal behavior.\\
    & Deep Learning for Anomaly Detection: A Review \cite{RF24} & Referred to as the process of detecting data instances that significantly deviate from the majority of data instances. \\
    & A Comprehensive Survey on Graph Anomaly Detection With Deep Learning \cite{RF25} & Anomalies might appear as social spammers or misinformation in social media; fraudsters, bot users or sexual predators in social networks; network intruders or malware in computer networks and broken devices or malfunctioning blocks in industry systems. \\
    \hline
    Outliers & Outlier Detection: Methods, Models, and Classification \cite{RF27} & In some cases in statistics and machine learning, outliers refer to those data instances (sometimes erroneous data points) that make it harder to fit the desired model.\\
    & An Introduction to Outlier Analysis \cite{RF23} & An outlier is a data point that is significantly different from the remaining data.\\
    & Outlier Detection: A Survey \cite{RF19} & Outliers, as defined earlier, are patterns in data that do not conform to a well-defined notion of normal behavior, or conform to a well-defined notion of outlying behavior, though it is typically easier to define the normal behavior.\\
    \hline
    Novelty & A review of novelty detection \cite{RF22} & Test data that differ in some respect from the data that are available during training.\\
    & Review of novelty detection methods \cite{RF28} & Novelty (anomaly, outlier, exception) is a pattern in the data that does not conform to the expected behavior.\\
    \hline
    Deviation & Group Deviation Detection Methods: A Survey \cite{RF21} & Group deviation detection involves the discovery of group behaviors that significantly deviate from the expected group patterns.\\
    \bottomrule
    \end{tabular*}
    \label{Table1}
\end{table}

\section{Relationship of different training methods}\label{3}

\subsection{Supervised learning}\label{3.1}

Supervised method means that the training set of the model must be explicitly labeled with normal and abnormal samples to learn the decision boundary, probability distribution or determine the confidence interval from the annotated instances. Since the dataset contains more prior knowledge, supervised models can achieve better performance than semi-supervised or unsupervised models, but the premise is to obtain large, explicitly labeled training data. As a result, the challenge of data imbalance has emerged as the primary obstacle in supervised learning, making it arduous to effectively differentiate between normal and abnormal data characterized by high-dimensional and intricate features. At present, there is very little research on applying Transformer to supervised anomaly detection tasks. 

Li et al. \cite{RF30} proposed GTF, an adaptive integration approach to anomaly detection for performance-constrained network edge computing devices by combining TabTransformer \cite{RF31} and Gradient Boost Decision Tree (GBDT). They mitigated the imbalanced classification problem by $Focus Loss$ ($FL$) and extended $FL$ to multi-class classification tasks to achieve a total of 9 different types of network anomaly detection including dos attacks, worms, backdoors, etc. Experiments show that the supervised paradigm of GTF can improve the robustness of handling class imbalance data. Similar theories and ideas also appear in the work of Xu et al \cite{RF32}. CATLog \cite{RF33} introduced a contrastive learning approach to generate embedding layer perturbations based on the standard supervised learning process, and further enhanced the robustness of the model through entropy enhancement. NeuralLog \cite{RF34} performed supervised binary anomaly detection on log sequences without using log parsing algorithms. The author argued that although NeuralLog is a supervised method, it can be easily applied to semi-supervised scenarios, by simply using reconstruction loss instead of classification loss. 

Stanislav et al. \cite{RF35} investigated the application of Transformer in OOD detection tasks, considering diverse data patterns. They proposed a two-step approach, beginning with unsupervised pre-training of the model followed by fine-tuning in a supervised setting. Their method surpasses previous SOTA models in OOD benchmark tests, demonstrating its efficacy. OODformer \cite{RF36}, uses ViT with DeiT \cite{RF37} to detect the OOD problem of image multi-class classification tasks in a supervised manner. Yu et al. \cite{RF38} used marked data from private power grids and trained BERT model to detect Advanced Persistent Threats (APT) attack sequences for the industrial Internet. Similar to OODformer, they also carried out the multi-class classification detection task by classifying simulated attacks into five categories (Normal, Detection, DOS, U2R and R2L), feeding the data into a pre-trained BERT model to extract feature representations related to APT attacks, followed by multi-class classification training through a $Softmax$ classifier. SwissLog \cite{RF39} is also a supervised binary log anomaly detection model, which combines semantic embedding and temporal embedding to detect faults resulting in sequence order changes as sequential log anomalies and time interval changes as performance issues. Guo et al. \cite{RF40} used the stacked Informer \cite{RF41} to extract underlying features of long time series data in a supervised way, and then replaced the original Adam optimizer with the method of Gradient Centralized (GC) + optimizer for anomaly detection of power line tripping faults.

In general, supervised methods have higher application value in multi-class classification anomaly detection tasks. Moreover, it can be seen from related works that supervised approaches are also gradually merging with unsupervised and semi-supervised paradigms toward universality and robustness.

\subsection{Semi-supervised learning}\label{3.2}

Semi-supervised learning is suitable for training tasks with large amounts of data and high labeling costs. The usual approach of a semi-supervised paradigm is to train the deep learning network by labeling only a portion of training data. For anomaly detection tasks, semi-supervised learning methods typically involve training the model by adding only normal samples to the dataset. With partial data labeling, semi-supervised learning can usually achieve better performance than unsupervised learning, but the features extracted from latent layers may not be the actual anomalies and are prone to overfitting problems. Manolache et al. \cite{RF42} proposed DATE, a Transformer-based model for text anomaly detection tasks. They tested both semi-supervised training and unsupervised training conditions in the experimental stage. Experimental results show that Transformer is highly generalizable and has excellent performance in a variety of test environments. Zhao et al. \cite{RF43} proposed Trine, a system log anomaly detection method using a semi-supervised paradigm. Trine uses only normal samples for training, adopts minimax strategies to train generators and discriminators, and performs anomaly detection on system logs. VT-ADL \cite{RF44} uses a similar paradigm for anomaly detection (image-level fine-grained) and anomaly location tasks (pixel-level fine-grained).

\subsection{Unsupervised learning}\label{3.3}

For anomaly detection, unsupervised learning usually refers to the case where it is difficult to obtain labels for anomalous data. The core of an unsupervised anomaly detection task is to distinguish between normal samples and abnormal samples based on the latent space of feature distribution. In this case, there is an underlying assumption that normal cases occur more frequently than anomalies, otherwise the model would have a high false positive rate. Therefore, in a sense, the unbalanced distribution of data can be considered a prerequisite for unsupervised anomaly detection. At present, the majority of studies exploring the application of Transformer in anomaly detection predominantly revolve around unsupervised models, with a primary emphasis on tackling binary classification problems. There is little research on using unsupervised methods to perform multi-class classification anomaly detection tasks. However, unsupervised paradigms are very sensitive to abnormal data noise, and face the challenge of learning the inherent commonalities of data in a complex high-dimensional space. Their performance is also inferior to supervised and semi-supervised methods. Therefore, for unsupervised learning methods, the priority is to ensure the stability of network training and to obtain higher anomaly detection accuracy. 

A2Log \cite{RF45} employs an unsupervised approach to establish optimal decision boundaries for the model. It initiates the anomaly scoring process on datasets such as BGL, thunderbird, and spirit1, followed by anomaly decision-making based on the obtained scores. Xu et al. \cite{RF46} believed that the key to unsupervised time series anomaly detection is to learn information representation and find distinguishable standards. They used Transformer to find more anomaly associations and solved the problem with association discrepancy. CT-D2GAN \cite{RF47} applied an unsupervised approach to the video anomaly detection task. The authors used a hybrid model based on convolutional Transformer and dual discriminator to perform anomaly detection on videos by extracting spatial information from the convolutional encoder and contextual temporal information from Transformer. Similar to CT-D2GAN, DCT-GAN \cite{RF48} is also a hybrid model based on Transformer, CNN, and GAN for unsupervised time series anomaly detection. Experimental results show that the training strategy of DCT-GAN with sliding window, GAN loss, and Gradient Penalty (GP) could enhance the robustness of the unsupervised model within a certain range of anomaly rates $ (5\%-20\%) $. Tajiri et al. \cite{RF49} used the structure of Set Transformer \cite{RF50} to detect anomalies in unsupervised information and communication technology systems (ICT). 

LAnoBERT \cite{RF51}, and LSADNET \cite{RF52} are both unsupervised anomaly detection models for logs. The distinction lies in the approaches employed by LAnoBERT and LSADNET. LAnoBERT utilizes the BERT model without relying on a log parser. It detects anomalies by leveraging a preprocessing technique that involves minimization and a regularization approach to handle unstructured text, such as log sequences. These preprocessed sequences are then inputted into the BERT model for anomaly detection through log sequence masking and prediction. While LSADNET draws inspiration from both Transformer and CNN techniques used in image anomaly detection. It divides the log anomaly detection task into two key aspects: local information extraction and global sparse Transformer. Local information extraction entails capturing local correlations through multi-layer convolution. In contrast, global sparse Transformer utilizes the Transformer model to learn global correlations among remote logs. GTA \cite{RF53}, MT-RVAE \cite{RF54} are unsupervised methods for anomaly detection of MTS data. GTA uses Transformer and Graph Convolutional Networks (GCN) to automatically learn graph structure and model temporal correlations. MT-RAVE, on the other hand, uses an up-sampling algorithm to capture multi-scale temporal information, employs a self-attention mechanism to capture potential correlations between sequences, and integrates the extracted features through a residual VAE.  

Metaformer \cite{RF55} is an unsupervised universal anomaly detection model combining Model-agnostic meta learning (MAML) and Transformer, which can solve the anomaly classification and anomaly location tasks in image anomaly detection. Schneider et al. \cite{RF56} evaluated with an unsupervised paradigm including Reconstruction Convolutional Autoencoder (R-CAE), Prediction Convolutional Autoencoder (P-CAE), Prediction Convolutional LSTM (P-ConvLSTM), Reconstruction Vision Transformer (R-ViT-AE) and Prediction Vision Transformer (P-VIT-AE), and investigated their performance for TOF depth image anomaly detection. 

Pinaya et al. \cite{RF5} integrated a vector-quantized VAE (VQ-VAE) and an autoregressive Transformer model to learn the brain's 2D image probability density function for unsupervised brain anomaly detection and segmentation tasks. Xu et al. \cite{RF57} employed a pre-trained Transformer model and fine-tuning techniques for unsupervised OOD detection. They utilized the Mahalanobis distance to extract potential representations from all layers of the pre-trained BERT model. By identifying the layers with more significant feature importance, they were able to reduce the problem to low-dimensional constrained convex optimization. This approach effectively streamlined the detection process. UTRAD \cite{RF58} is a U-Transformer model using the skip-connection method for unsupervised image anomaly detection. UTRAD firstly extracts multi-scale features through a pre-trained CNN trunk, then uses U-Transformer as the multi-scale reconstruction model for feature reconstruction, and adopts reconstruction error to measure the anomaly score. UCAD \cite{RF59} innovatively applies Transformer to the unsupervised contextual anomaly detection task for database systems. 

Although the unsupervised paradigm may not exhibit superior performance compared to other approaches, it is better aligned with the requirements of most anomaly detection tasks in unlabeled contexts. Additionally, this approach finds application in diverse fields, including images, logs, time series, etc. The detection performance can be further improved by different auxiliary methods such as data enhancement. Therefore, the unsupervised paradigm is one of the current research hotspots in anomaly detection.

\subsection{Self-supervised learning}\label{3.4}

Self-supervised learning is essentially a variant of unsupervised learning methods. Different from unsupervised learning, self-supervised learning mainly uses a pretext to mine its supervised information from large-scale unsupervised data and trains the network with the constructed supervised information, which in turn can learn representations that are valuable for downstream tasks. In other words, the labels of self-supervised learning are not manually labeled and can be trained using a similar approach to supervised learning after the labels are obtained through pretexts. 

Guo et al. \cite{RF16} used BERT to detect anomalies in log sequences by training two self-supervised tasks, namely masked log message prediction and hypersphere volume minimization. Marino et al. \cite{RF60} proposed Network Transformer, a network traffic anomaly detection framework using Transformer. Network Transformer draws lessons from the idea of Graph Neural Network (GNN), extracts the corresponding hierarchical graph features from the network nodes, and trains the neural network in a self-supervised way by collecting data only during the normal operation of the system. Park et al. \cite{RF61} adopted the UNETR model for self-supervised learning, intending to solve the inaccuracy of OOD detection caused by insufficient labeled data. They pointed out that some anomalous images of rare diseases in the medical field are very difficult to obtain, so self-supervised learning is of great significance in the research field of medical OOD detection. Guo et al. proposed TransLog \cite{RF62}, a self-supervised anomaly detection method composed of pre-training and adapter-based tuning phases. TransLog uses the method of transfer learning to improve the generalization ability of anomaly detection. Mai et al. \cite{RF63} conducted experiments to evaluate the detection performance of a fine-tuned Transformer model on near OOD tasks, specifically focusing on semantic and syntactic anomalies. Through a self-supervised approach, they confirmed the exceptional feature extraction capabilities of Transformer.

Although relevant research related to Transformer-based self-supervised anomaly detection is still in its infancy, it is foreseeable that the stronger generalization ability and model robustness make the self-supervised approach one of the high-potential paradigms.

\subsection{Weak-supervised learning}\label{3.5}

Weak-supervised learning is a new branch in the field of machine learning. Compared with supervised methods, weak-supervised learning trains model parameters by using limited, noisy, or inaccurately labeled data. According to the degree of data labeling, the weak-supervised learning method can be divided into the following three categories: incomplete supervision, inexact supervision, and inaccurate supervision. Incomplete supervision means that part of the data in the training sample contains labels, while the rest do not have valid labels, similar to semi-supervised learning. Therefore, in a sense, weak-supervised learning and semi-supervised learning are not completely mutually exclusive. Inexact supervision means that the data has only coarse-grained labels. Inexact supervision is most often applied in the image domains, such as the task of only giving information about the presence of a face in an image, but not providing the exact location of the face. For anomaly detection tasks, inexact supervision often means that the coarse-grained information is labeled with anomalies (i.e., giving whether there is an anomaly in the entire time series or the whole image), and the model needs to perform the anomaly detection task of fine-grained information (i.e., determining whether there is an anomaly at a certain time step in the time series data or in a certain pixel position of an image). Inaccurate supervision means that the training samples contain labels but are not accurate. The reason for this phenomenon may be that the data contains noise, human mislabeling, etc. 

Li et al. \cite{RF64} used a Convolutional Transformer-based Multi-Sequence Learning network for video anomaly detection from two different granularities (video-level anomaly probability and fragment-level anomaly score). They used whether the video-level fine-grained contains anomalies as a label to predict the anomaly score of each frame through a weak supervision paradigm, which is a typical inaccurate supervision approach. Huang et al. \cite{RF65} proposed a weakly supervised temporal discriminative (WSTD) paradigm to address this problem from another perspective. In addition, Tian et al. \cite{RF66} used a weak-supervised learning method of Convolutional Transformer to detect the anomalies of polyp frames.

Similar to self-supervised learning, the development of weak-supervised paradigm is also in its infancy. However, existing experimental results \cite{RF65} demonstrate that weak-supervised approaches can achieve improved anomaly detection performance compared to unsupervised methods while providing easier access to label information compared to fully supervised methods. As a result, weak supervision is poised to become one of the prominent research focal points in this field.

\begin{figure}[H]
    \centering
    \includegraphics[width=1.0\linewidth]{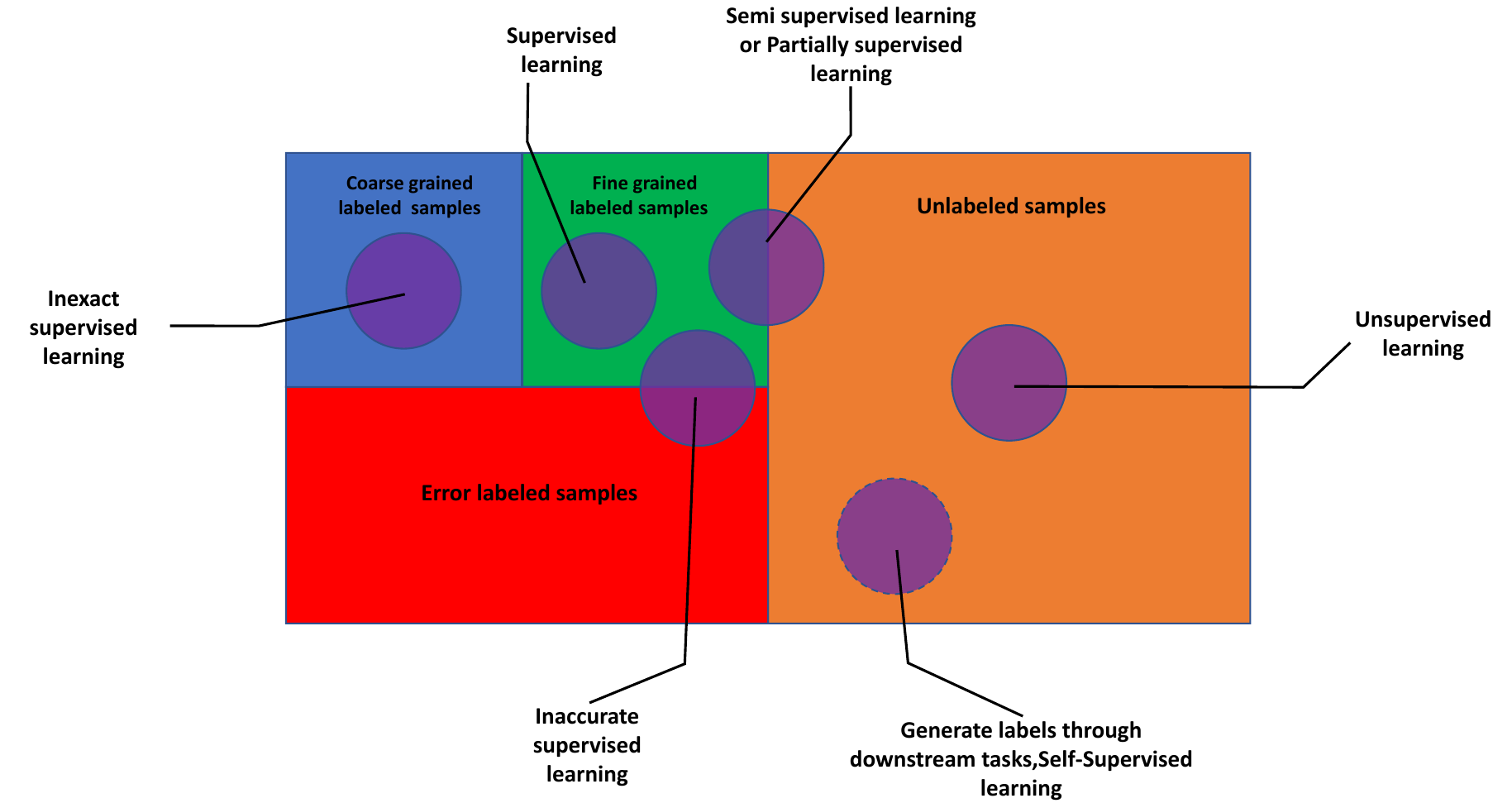}
    \caption{The relationship diagram of supervised, semi-supervised, unsupervised, self-supervised, and weak-supervised learning}
    \label{figure1}
\end{figure}

Finally, the structural relationships of various paradigms are shown in Figure \ref{figure1}.

\section{Research of Transformer based anomaly detection} \label{4}

\begin{figure*}[h]
    \centering
    \includegraphics[width=0.76\textwidth]{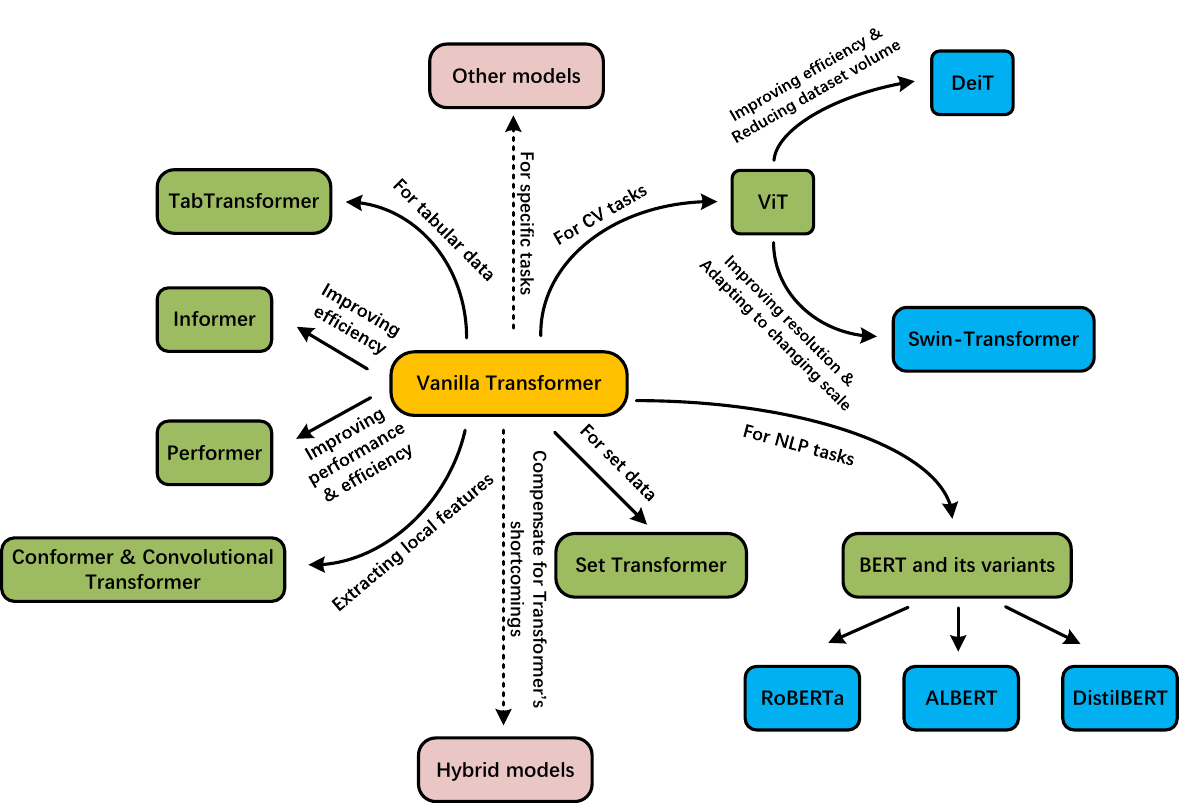}
    \caption{Relationship between different Transformer variants}
    \label{figure2}
\end{figure*}

In this paper, we divide the research in this area into anomaly detection based on Vanilla Transformer, anomaly detection based on Transformer variants, and anomaly detection based on hybrid methods. (Note: Hybrid methods refer to the combination of Transformer or attention mechanism with other methods, such as Transformer + GAN, Transformer + VAE, attention + LSTM, etc.) The relationship between various Transformer variants is shown in Figure \ref{figure2}. We provide a concise overview of the operation principle of each Transformer variant model, analyzing their respective advantages and shortcomings.

\subsection{Anomaly detection based on Vanilla Transformer} \label{4.1}
Google first proposed the Vanilla Transformer \cite{RF14} in 2017 as the basis for many subsequent variants, whose core is mainly composed of Position Encoding, Multi-head Attention (MHA) mechanism, Self-Attention Layer, Feed-Forward and Residual Network.
~\\

\textbf{Position Encoding}

Since Transformer does not have a linear input/output structure like LSTM and RNN neural networks, Position Encoding is introduced in Transformer to ensure the correctness of position information. In the Vanilla Transformer, Position Encoding information is achieved by $\sin$ and $\cos$ function, as shown in equation \eqref{eq:1}:
\begin{equation}
    \label{eq:1}
    \Vec{p_t}=f(t)^{(i)}=\begin{cases}
    \sin{(\omega_k\cdot t)}, & \text{if } (i = 2k);\\
    \cos{(\omega_k\cdot t)}, & \text{if } (i = 2k+1).
    \end{cases}
\end{equation}

After the Position Encoding is calculated by the above equation, it also needs to be added to the model input, as shown in equation \eqref{eq:2}:

\begin{equation}
    \label{eq:2}
    \phi'(\omega_t)=\phi(\omega_t)+\vec{p_t}
\end{equation}

Since it is necessary to ensure correct vector addition, the dimension of the position vector $\vec{p_t}$ must be consistent with the input dimension.
~\\

\textbf{MHA}\label{4.1.2}

For each attention head, Query ($Q$), Key ($K$) and Value ($V$) matrices are learnable parameters. Therefore, the scaled dot-product attention defined by Transformer is shown in equation \eqref{eq:3}:

\begin{equation}
    \label{eq:3}
    Attention(Q,K,V)=softmax(\dfrac{QK^T}{\sqrt{D_k}})V
\end{equation}

Where matrix $ Q \in R^{N \times D_k} $,$ K \in R^{M \times D_k} $,$ V \in R^{M \times D_v} $. $N$,$M$ represents the length of $Q$,$K$, and $ D_k $,$ D_V $ represents the dimension of matrix $K$,$V$. The $Softmax$ function is used to compress the operation results into a smaller feature embedding space. Researchers have found that applying multiple attention heads simultaneously to capture different features has better performance \cite{RF14}. Thus, the calculation method of MHA is shown in equation \eqref{eq:4}:

\begin{equation}
    \label{eq:4}
    MHA(Q,K,V)=Concat(head_1,head_2,...,head_H)W^O 
\end{equation}

Where $ head_i=Attention(QW_i^Q, KW_i^K, VW_i^V) $, $MHA$ represents multi-head attention.
~\\ 

\textbf{Self-Attention Layer}

The Self-Attention layer applies the MHA mechanism described in section \ref{4.1.2} to learn the corresponding long-term dependencies. A key factor affecting the ability to learn such dependencies is the path length that forward and backward signals must traverse in the network. The shorter the path between any combination of positions in the input and output sequence, the easier it is to understand the long-term dependencies \cite{RF14}. LSTM and GRU neural networks lose some memory information when faced with excessively long input sequences, while Transformer's self-attention layer achieves better performance than LSTM and GRU through the MHA mechanism.
~\\

\textbf{Feed-forward and Residual Network}

The feed-forward neural network is a neural network with a fully connected structure, which is used to transfer information to the next neural network layer, as shown in equation \eqref{eq:5}:

\begin{equation}
    \label{eq:5}
    FFN(H')=ReLU(H'W^1+b^1)W^2+b^2
\end{equation}

Where $H'$ is the output of the previous neural network layer, $FFN$ refers to the Feed-forward neural network. $ W^1 \in R^{D_m \times D_f} $, $ W^2 \in R^{D_f \times D_m} $, $ b^1 \in R^{D_f} $, $ b^2 \in R^{D_m} $ are all trainable parameters. In the deeper model, to prevent the information bottleneck, Transformer applies the residual connection modules, as shown in equation \eqref{eq:6} and \eqref{eq:7}:

\begin{equation}
    \label{eq:6}
    H'=LN(SelfAttn(X)+X)
\end{equation}

\begin{equation}
    \label{eq:7}
    H=LN(FFN(H')+H')
\end{equation}

Where $SelfAttn$ represents the self-attention layer mentioned in section \ref{4.1}, and $LN$ represents the operation of the normalization layer.

The core innovation of Vanilla Transformer lies in the excellent feature extraction capability and parallel computing feature of the MHA mechanism, so researchers often apply it to anomaly detection tasks of serialized data, including but not limited to time series data, log data, stream data, etc. 

Kozik et al. \cite{RF67} used Vanilla Transformer to detect anomalies in network traffic generated by IoT devices. They collected data in the form of stream data collection on the Aposemat IoT-23 dataset and converted it to a sliding window format as the input of Transformer. Zhang et al. \cite{RF3} applied Vanilla Transformer to an ICS traffic anomaly detection task and obtained the final anomaly detection result through a classifier. A2Log \cite{RF45} only uses the Transformer encoder structure, and performs the task of anomaly scoring and anomaly decision on this basis. A2Log's experiments demonstrate the universality of Vanilla Transformer, which can achieve excellent performance on different log datasets. Similarly, Li et al. \cite{RF68} used only Transformer encoder to extract the hidden features of traffic and perform anomaly detection. Falt et al. \cite{RF69} argued that training a single Transformer model would suffer from large performance differences, so they used a bagging strategy to ensemble many Transformers, where each underlying Transformer votes on the prediction results. The system selects the highest prediction score for anomaly detection of log sequences by calculating the number of predicted votes and multiplying them by the corresponding weights. However, the performance overhead of this method is very large, as well as low training efficiency. 

Meng et al. \cite{RF6, RF70} used the structure of Vanilla Transformer for anomaly detection of spacecraft datasets (MSL, SMAP). UT-ATD \cite{RF71} uses Vanilla Transformer to learn the embedding of trajectory mapping and detect anomalous trajectory from trajectory embedding through a Multilayer Perceptron (MLP) layer. Huang et al. \cite{RF2} proposed HitAnomaly, a model for log anomaly detection using a hierarchical Transformer structure and pooling layer. They used a log encoder and a parameter value encoder to encode log templates and parameter values respectively, and used an attention-based approach to classify anomaly detection results. Similarly, Xiao et al. \cite{10138083} also applied Vanilla Transformer to log anomaly detection tasks and designed a more flexible and robust 'top-p' algorithm. Zhang et al. \cite{9774889} designed a Transformer-enabled feature encoder to convert the input task-agnostic features into discriminative task-specific features by mining the semantic correlation and position relation between video snippets. Wibsono et al. \cite{RF72}, and Guo et al. \cite{RF73} both applied Vanilla Transformer only to the task of anomaly detection on log data. They did not propose too many modifications and thus had limited results.

As the research progresses, the drawbacks of Vanilla Transformer become apparent. The performance overhead of Vanilla Transformer is too large to be deployed in edge computing devices. For example, the adaptive Spatio-Temporal Attention Transformer proposed by Kumar et al \cite{SIVAKUMAR2023100625}. requires the simultaneous computation of temporal attention and MHA, resulting in significant performance overhead. Further, the inherent encoder-decoder structure greatly limits the application scenarios of Vanilla Transformer, so the Transformer decoder has to be discarded for better transfer learning tasks \cite{RF45}. Therefore, Ding et al \cite{ding2023concept}. made improvements to the Vanilla Transformer. They proposed the concept drift adaptation method (CDAM), a kind of distribution adaptation method, to dynamically tune the learning rate of the Transformer. They also utilized root square sparse self-attention, which only requires $O(L\sqrt{L})$ time complexity, for anomaly detection in time series data.

\subsection{Anomaly detection based on ViT}

The origin of ViT \cite{RF17} is to accomplish tasks in the field of Computer Vision (CV) using only the attention mechanism in Vanilla Transformer through transfer learning, without relying on CNN structure. ViT essentially divides the original images into blocks, flattens them into sequences, feeds them into the Encoder of Vanilla Transformer, and finally classifies the images through a fully connected layer. Specifically, suppose the original input image data is $ H \times W \times C $, Where $H$, $W$, and $C$ represent the length, width, and number of channels of the image, respectively. Then ViT first divides the picture into $N$ blocks, where $ N = (H \times W)/(P \times P) $, then each block is flattened into a 1-dimensional vector with vector size of $ P \times P \times C $. Therefore, the total input is then transformed into $ n \times (P^2 \times C) $. Then, a linear transformation is performed on each of the input vectors, and the dimension is compressed to $D$. ViT adds an extra dimension based on the Position Encoding of the Vanilla Transformer. The reason is that ViT only uses the encoder structure, so the added dimension is an extra dimension for classification. It is also a learnable parameter, which is concatenated with input. Many ViT-based anomaly detection models also use this dimension to classify anomalies.

Yu et al. \cite{RF74} used ViT and ResNet as feature extractors to extract global features and local patches. Then, the extracted features are passed through the FastFlow model proposed in this paper for anomaly detection and anomaly location. The essence of FastFlow is to construct a bijective invertible mapping that projects image features into hidden vectors. To preserve the spatial positional relationship within the feature map and prevent the flattening and compression of visual features from a two-dimensional space into a one-dimensional space, which could lead to the loss of crucial information, FastFlow incorporates a subnet equipped with a two-dimensional convolution layer in its flow model. This ensures that the corresponding spatial information is retained effectively. Stanislav et al. \cite{RF35} experimentally demonstrated on CIFAR-100, and CIFAR-10 datasets that the pre-trained ViT can achieve excellent performance in near OOD tasks. They also pointed out that pre-trained ViT is robust to input disturbances. 

Wrust et al. \cite{RF75} investigated the application of ViT in the novelty detection of traffic scenario infrastructure. They performed fine-tuning on the output of Vanilla ViT by transforming it from predictive class labels to latent representations. Additionally, they employed ViT within a triplet loss-based Autoencoder framework, generating latent representations for input road infrastructure images using ViT. Through metric learning, they aimed to move the latent representations of negative examples away from the baseline and closer to the latent representations of positive samples. They applied this model structure to many existing Novelty detection methods, such as Local Outlier Factor (LOF), Isolated Forest, Angle-based Outlier Detection, One-class Support Vector Machine (OCSVM), and UMAP-based Local Entropy Factor, with good experimental results. The UNETR model used by Park et al. \cite{RF61} is essentially a 3D UNET model, in which the encoder part uses the ViT structure. They converted the input images into sequence representations and connected them to the decoder utilizing a skip-connection structure. 

TSViT \cite{RF76} is an improved ViT model for time series tasks. They performed the anomaly detection task for the video tracking problem by encoding frames with only pedestrian position information into the embedded token and passing the spatial information of the pedestrian to TSViT. The author classified the detection features of TSViT into low-level features (3D gradient features, optical flow multi-scale histograms, structural context descriptors, and dynamic texture blending) and high-level features (object trajectory features, object appearance features). ViTALnet \cite{10054622} first employs ViT to extract local discriminatory features as feature representations. Subsequently, it introduces an anomaly estimation module that integrates global attention and a pyramidal architecture to enhance contextual information for fine-grained anomaly localization. VT-ADL \cite{RF44}, on the other hand, applies ViT to encode the input image that has been subdivided into small pieces, and subsequently feeds the ViT-encoded image to the decoder for image reconstruction, forcing the network to learn the features representing the normal image, thus performs the image anomaly detection task in a reconstruction manner. VT-ADL additionally introduces a Gaussian mixture density function for the pixel-level image anomaly location. Gaussian mixture density function models the distribution of ViT-encoded features to estimate the distribution of normal data in the potential space. To achieve this goal, VT-ADL slightly modifies the structure of ViT by discarding the dropout layer in the network, which would lead to the instability of the Gaussian approximation network, and adds a MLP layer, which is a linear layer containing two GELU activation functions. Thus, VT-ADL can simultaneously perform image anomaly detection and anomaly location tasks from different fine-grained levels. 

Fan et al. \cite{RF77, fan2023transformer} introduced a contrastive learning paradigm for ViT, using ViT as a feature extractor and performing image anomaly detection step-by-step through a contrast learning framework to mitigate the catastrophic forgetting problem. Lin et al. \cite{LIN2022104544} utilized ViT-S self-supervised learning to reconstruct unlabeled pavement images. They proposed an encoding-retrieval-matching pavement anomaly detection method to address the classification retraining problem. De Nardin et al. \cite{de2022masked} adopted a masked multi-head self-attention mechanism that allows the model to learn a relationship between different patches of the input images. They modified the architecture of ViT to achieve high-precision image anomaly detection by adding a new masking component and calculating attention between patches of different shapes. Lee et al. \cite{9956507} utilized the ViT model to perform video anomaly detection tasks from the spatio-temporal context perspective. They focused on 3 different contextual prediction streams: masked, whole, and partial.

Although ViT has been widely used in various image and video anomaly detection tasks, it still has many limitations. Vanilla Transformer was originally applied to NLP tasks, but the amount of information contained in CV tasks has increased by orders of magnitude compared to NLP tasks (e.g., an 8-megapixel color image contains far more pixel points than a text paragraph consisting of 200 words), so ViT's "brute force" processing method undoubtedly amplifies the disadvantage of Transformer's performance overhead. In addition, ViT usually processes low-resolution images (e.g., mosaic images with a resolution of $328 \times 328$) and often requires longer training time to achieve slightly better performance than CNN. Therefore, we believe that research related to anomaly detection based on ViT must solve the above problems to have a high practical application value.

\subsection{Anomaly detection based on Data-efficient image Transformer (DeiT)}

\begin{figure}[H]
    \centering
    \includegraphics[width=0.55\linewidth]{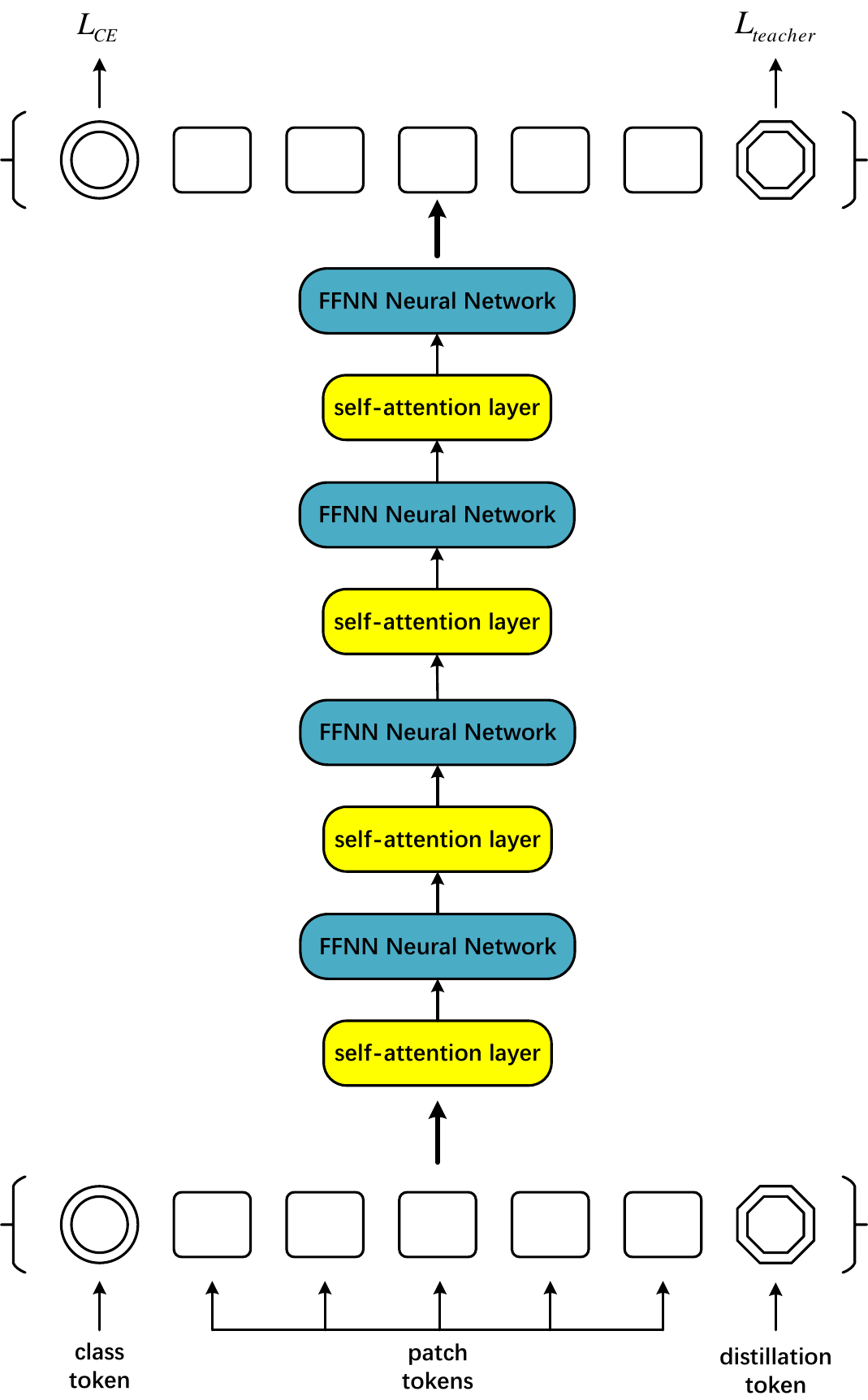}
    \caption{The structure of DeiT model}
    \label{figure3}
\end{figure}

DeiT \cite{RF37} is further improved based on ViT. Researchers have discovered that ViT demonstrates superior performance compared to traditional neural network architectures like CNN, but primarily on extremely large datasets, such as those consisting of 300 million images. This observation highlights the requirement of a substantial dataset for Transformers to achieve optimal performance and generalization capabilities. Consequently, this limitation hinders the broader application and advancement of ViT. In contrast, DeiT adopts a better training strategy based on ViT and employs distillation operation to achieve stable training performance while significantly reducing the dataset volume and training time. The structure of DeiT is shown in Figure \ref{figure3}.

As can be seen from Figure \ref{figure3}, DeiT adds a Distillation token based on ViT, which interacts with the class token and patch token of the original ViT in each multi-head self-attention (MSA) layer of the Transformer. However, the goal of the class token in ViT is to be consistent with the ground truth label, while the goal of the distillation token is to be consistent with the label predicted by the teacher model. Here, the teacher model is a high-performance classifier, such as a trained CNN model or Transformer model.

The distillation loss in DeiT can be divided into two categories, which are soft distillation loss and hard distillation loss. The equation for soft distillation is as follows:

\begin{equation}
    \label{eq:8}
    L^{softDistill}_{global}=(1-\lambda)L_{CE}(\psi{(Z_s)},y)+\lambda\tau^2KL(\psi{(\dfrac{Z_s}{\tau})},\psi{(\dfrac{Z_t}{\tau})})
\end{equation}

The equation for hard distillation is as follows:

\begin{equation}
    \label{eq:9}
    L^{hardDistill}_{global}=\dfrac{1}{2}L_{CE}(\psi{(Z_s)},y)+\dfrac{1}{2}L_{CE}(\psi{(Z_s)},y_t)
\end{equation}

Where $y_t=argmax_cZ_t(c)$.
$Z_s$ and $Z_t$ are the outputs of the student model and teacher model respectively,$KL$ stands for KL divergence, $\psi$ is $softmax$ function, $\lambda$ and $\tau$ are hyperparameters, and $CE$ denotes cross entropy. 

Hence, it can be deduced that in the case of soft distillation, the output $Z_s$ of the student network is compared with the true label using the $CE_{Loss}$ function to obtain the intermediate result $O_1$. This result is then further compared with the $softmax$ output $O_2$ of the teacher network using the $KL_{Loss}$ function, resulting in the final calculated output $O_3$ with an additional weight averaging operation. On the other hand, in the case of hard distillation, the output $Z_s$ of the student network is compared with the true label using the $CE_{Loss}$ function to obtain the intermediate result $O_1$. Subsequently, $Z_s$ is compared with the output of the teacher network using the same $CE_{Loss}$ function, yielding the intermediate result $O_2$. Finally, an average operation is performed on these intermediate results to obtain the final output $O_3$.

Since DeiT adds an extra distillation token compared to ViT, the objective function of its output is exactly the soft distillation loss function or hard distillation loss function mentioned above, Therefore, DeiT adds the output vector of the class token and the output vector of the distillation token through the linear layer to obtain the final prediction result after $softmax$ addition during the test.

OODformer \cite{RF36} uses a stronger data enhancement method with the ImageNet-21K dataset to pre-train DeiT, and uses embedding similarity methods to judge outliers. The authors concluded that under ideal circumstances, the representational similarity of OOD samples is much lower than that of ID samples, and their $softmax$ confidence should be significantly lower than that of ID samples. Finally, the author used two methods to judge OOD samples, which are the distance measure of potential spatial embedding and the $softmax$ confidence score. However, this paper also pointed out that the application of DeiT in OOD detection tasks is still in its infancy and there is no universal framework. The experiment in this paper is only an early attempt at ViT and DeiT models for the OOD detection task, which needs to be further studied and explored. 

Thus, although DeiT addresses some of ViT's pain points, significant data enhancement is still needed to expand the training set capacity and achieve better anomaly detection performance.

\subsection{Anomaly detection based on TabTransformer}

TabTransformer \cite{RF31} is a model for processing tabular data, which treats each row in a table as a sentence and each column value as a word or token. Since the length of each row in the table is fixed, and the position of each token in the table is relatively fixed, TabTransformer discards the Position Encoding method of Vanilla Transformer and only needs to focus on fixed-length input, which is shown in Figure \ref{figure4}:

\begin{figure}[H]
    \centering
    \includegraphics[width=0.6\linewidth]{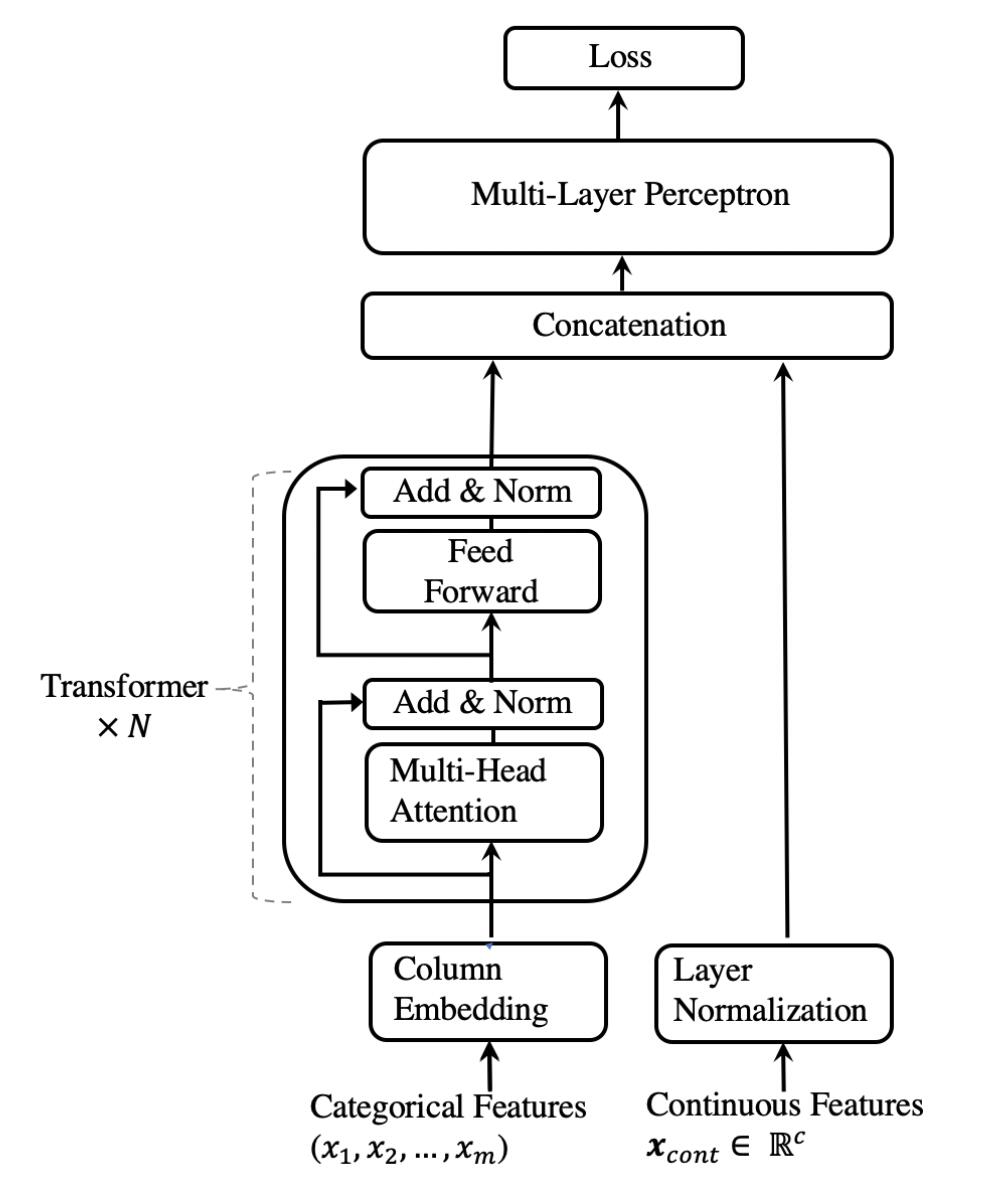}
    \caption{The structure of TabTransformer (NOTE: The figure is from the paper \cite{RF31})}
    \label{figure4}
\end{figure}

The main innovation of TabTransformer is its ability to train for semi-supervised scenarios effectively and has relative robustness to handle missing and noisy data. However, it is also easy to see from Figure \ref{figure4} that TabTransformer also has significant structural defects. Since TabTransformer only sends categorical features into the Transformer Block for feature extraction after vector encoding, continuous features are only connected to the output of the Transformer block after a simple normalization operation. Therefore, if the input contains few category features and many continuous features, the structure of TabTransformer can bring very limited gain. On the contrary, the training efficiency will be reduced due to too many Transformer training parameters. Therefore, most of the TabTransformer-based anomaly detection models use the ensemble learning method to compensate for the deficiencies of TabTransformer. For example, GTF \cite{RF30} simplifies TabTransformer by removing the continuous features in the input, as well as the normalization layer and connection layer, to take full advantage of the TabTransformer's ability in processing tabular input data. GTF also converts GBDT into a neural network structure GBDT2NN using the expressive ability of neural networks, to better integrate with TabTransformer. To enhance the training efficiency of the ensemble method, GTF incorporates the adaptive training technique. This approach defines a range of parameter values to be trained by setting upper and lower limits. Additionally, a target index, such as the $F1$ score, is specified to automatically search for the optimal parameter configuration that yields the best performance during the training process. By employing this adaptive training method, GTF aims to optimize its performance effectively. All the above optimization measures ensure that GTF retains the performance of TabTransformer while reducing the computational complexity required for model training, allowing GTF to be deployed on performance-constrained edge computing devices.

\subsection{Anomaly detection based on Swin-Transformer}

Swin-Transformer \cite{RF79} proposed by Microsoft is a universal Transformer architecture for CV tasks. Similar to DeiT, it is improved based on ViT and is mainly used to solve 2 major problems of Transformer in CV tasks:

\begin{enumerate}
    \item In NLP tasks, a Token is used as the input, so the scale is relatively fixed, but the scale varies greatly in CV tasks;
    \item CV requires higher processing resolution (higher fine-grained) than NLP tasks, and the computational complexity is often the square of NLP tasks. Transformer has difficulty adapting to pixel-level tasks on high-resolution images.
\end{enumerate}

\begin{figure*}[h]
    \centering
    \includegraphics[width=0.97\textwidth]{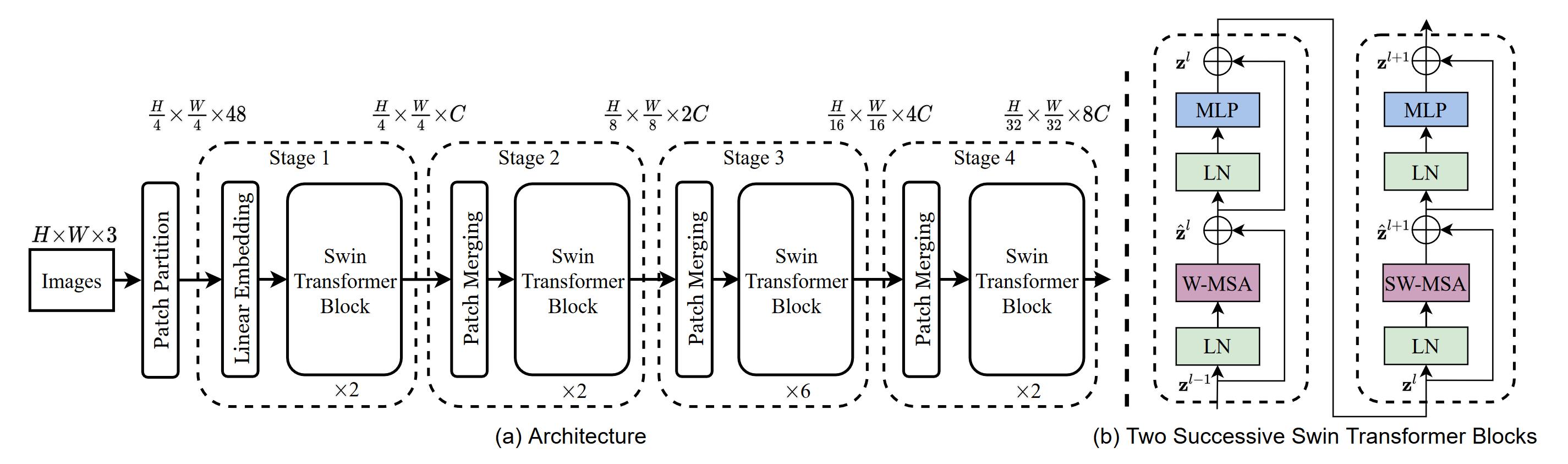}
    \caption{The structure of Swin-Transformer(NOTE: The figure is from the paper \cite{RF79})}
    \label{figure5}
\end{figure*}

The core structure of Swin-Transformer is shown in Figure \ref{figure5}.

According to Figure \ref{figure5} (a), the processing flow of Swin-Transformer is as follows:

\begin{itemize}
    \item Step 1: Image preprocessing (blocking and dimension reduction)\hfill
    
    Swin-Transformer first converts an $H \times W \times 3$-size image into a 2D image patches of $x_P \in N \times (P^2 \times C)$. Therefore, the number of blocks is $ N=(H \times W)/P^2$, and the dimension of each block is $P^2 \times 3$, Where $P$ is the corresponding block size. In Swin-Transformer, $P=4$. Thus Swin-Transformer transforms an $H \times W \times 3$ image into a tensor of $H/4 \times W/4 \times 48$.
    
    \item Step 2: Stage 1\hfill
    
    In Stage 1, similar to ViT, a linear embedding is done first to transform 48 dimensions into C dimensions. So the dimension of the tensor becomes $h/4 \times W/4 \times C$. The tensor is then fed into the Swin-Transformer block, whose structure is shown in Figure \ref{figure5}(b). The first Swin-Transformer block is composed of a window-based MSA with two MLP layers, and the other is composed of a shifted window-based MSA with two MLP layers. It can be seen that the biggest improvement of the Swin-Transformer block compared with the Transformer block in ViT is the replacement of MSA with W-MSA and SW-MSA. MSA in ViT uses a global attention mechanism, which will greatly increase the computational cost in high-resolution images. W-MSA computes self-attention in each window. Assuming that each window contains $M \times M$ image patches, the computational volume of MSA and W-MSA can be expressed as:
    
\begin{equation}
    \label{eq:10}
    \Omega (MSA)=4h \omega C^2+2{(h \omega)}^2C
\end{equation}

\begin{equation}
    \label{eq:11}
    \Omega (W-MSA)=4h \omega C^2+2M^2h \omega C
\end{equation}
    
    Although W-MSA reduces the computation, it only has a local receptive field of window size, not the global receptive field of the whole image. Hence, SW-MSA is specifically designed to address this issue. It accomplishes this by employing an attention mechanism that establishes connections between different windows, enabling information exchange. Additionally, SW-MSA introduces connections between adjacent non-coincidence windows in the upper layer through a shifting division of windows. This approach significantly enhances the receptive field, allowing for a broader capture of contextual information. Meanwhile, W-MSA further ensures the constant number of windows by the cycle shift method. Therefore, the computational flow expression of the whole Swin-Transformer Block is as follows:

\begin{equation}
    \label{eq:12}
    \hat{z}^l=W-MSA(LN(z^{l-1}))+z^{l-1}
\end{equation}

\begin{equation}
    \label{eq:13}
    z^l=MLP(LN(\hat{z}^l))+\hat{z}^l
\end{equation}

\begin{equation}
    \label{eq:14}
     \hat{z}^{l+1}=SW-MSA(LN(z^l))+z^l
\end{equation}

\begin{equation}
    \label{eq:15}
     z^{l+1}=MLP(LN(\hat{z}^{l+1}))+\hat{z}^{l+1}
\end{equation}

    \item Step 3: Stage2/3/4\hfill
    
    The basic operations of Stage 2/3/4 are similar. At the beginning of Stage 2 - Stage 4, a patch merging operation is performed to reduce the number of tokens, and the adjacent $2 \times 2$ tokens are merged. Each stage changes the dimensionality of the tensor, forming a hierarchical representation that can be conveniently replaced by a backbone network for various visual tasks.
\end{itemize}

Swin-Transformer also further improves the Position Encoding method by superimposing the Position Encoding on the attention matrix instead of the $sin$ and $cos$ Position Encoding function of the Vanilla Transformer.

Due to the excellent performance of Swin-Transformer in CV tasks, it has been applied to anomaly detection tasks in CV domains such as image and video. Han et al. \cite{RF4} used the Swin-Transformer architecture with GP to extract image features. They further modified the loss function based on Swin-Transformer by replacing it with a loss function combining Negative Log Likelihood (NLL) and GP. Their work has strongly demonstrated the excellent performance of Transformer architecture for Autonomous Driving Systems (ADS) scene-aware anomaly detection. Swin-Transformer outperforms traditional single-class classifiers based on kernel SVM and machine learning classifiers based on Decision Trees on several datasets with different subtasks in the autonomous driving fields (Baidu Apollo for GPS spoofing attacks, GTSRB for traffic sign attacks, Tusimple for detecting lane attacks). On this basis, Han et al. \cite{RF80} added federated learning and life-long learning to Swin-Transformer and named the system ADS-Lead. ADS-Lead can continuously collect data and update models in real-time, to better perform online anomaly monitoring and anomaly detection tasks. Concretely, ADS-Lead’s federated learning technology communicates from each vehicle’s C-ITS system to the server cluster using V2C technology. The server receives multiple vehicle information and aggregates it into a gradient vector, then averages it for calculation, returning the new model to each vehicle in the C-ITS system. Each vehicle can use the latest system for anomaly detection, while the entire system is updated asynchronously, meaning that the system does not need to wait for all the information from each vehicle to be in place before calculation, but can receive information, for real-time calculation and real-time feedback. Jiang et al. \cite{9858596} proposed the use of Masked Swin Transformer Unet for image anomaly detection tasks. They modified the method of determining anomalies from image reconstruction to image inpainting and utilized the powerful global learning ability of Swin Transformer to inpaint the masked area. By dividing the input image into nonoverlapping image patches, putting the image patches into a Swin Transformer-based encoder to extract global contextual features, dual upsampling them with the decoder, and fusing with multiscale features from the encoder via skip connections, they generated masked patches and achieved high detection accuracy.

\subsection{Anomaly detection based on Informer}

Informer \cite{RF41} aims to improve the operational efficiency of the Vanilla Transformer and reduce the time complexity. Compared with Vanilla Transformer, the time complexity of Informer is only $O(L \log L)$, which makes the Informer competent for Long Sequence time-series Forecasting (LSTF) tasks. The main innovation of Informer lies in its proposed ProbSparse Attention Mechanism, self-attention distillation technique, and generative decoder to improve the efficiency of Transformer, whose overall structure is shown in Figure \ref{figure6}:

\begin{figure}[H]
    \centering
    \includegraphics[width=0.78\linewidth]{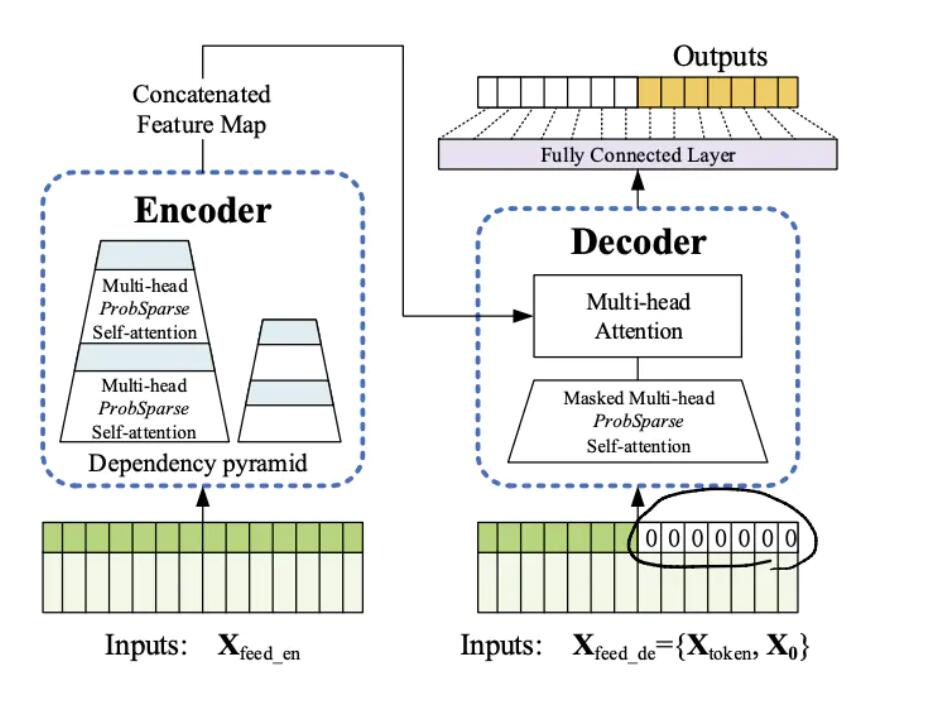}
    \caption{The overall structure of Informer (NOTE: The figure is from the paper \cite{RF41})}
    \label{figure6}
\end{figure}

\begin{itemize}
    \item Probsparse attention mechanism\hfill
    
    Before Informer, some theoretical studies revealed the potential sparsity of self-attention probability. However, Informer experimentally verified the sparsity of the original attention mechanism through experiments, i.e., the long-tail phenomenon of the attention feature map. Since only a small fraction of dot products contribute to the attention score, only the important part needs to be screened out for calculation. The author argued that the prominent dot product causes the attention probability distribution of the $Query$ matrix to be far away from the uniform distribution, so they used $KL$ divergence to measure the distance between the two distributions. The divergence measurement equation for the $i^{th}$ Query is as follows:

\begin{equation}
    \label{eq:16}
     M(q_i,K)=ln\sum_{j=1}^{L_K}e^{\dfrac{q_ik_j^T}{\sqrt{d}}}-\dfrac{1}{L_K}\sum_{j=1}^{L_K}\dfrac{q_ik_j^T}{\sqrt{d}}
\end{equation}

    If the $i^{th}$ query has a larger value $M$, its attention probability $p$ varies from other parts and, therefore has a high probability of being an important part. Thus, the final ProbSparse self-attention calculation equation is as follows:
    
\begin{equation}
    \label{eq:17}
     A(Q,K,V)=Softmax(\dfrac{\overline{Q}K^T}{\sqrt{d}})V
\end{equation}

Where $Q$ is a sparse matrix containing top U queries ($ u=c\times \log \ln LQ $, where $C$ is the control factor)

    \item Self-attention distillation technology\hfill
    
    The self-attention distillation can reduce the network parameters and "distill" the outstanding features as the number of stacked layers increases. The operational distillation equation from layer $J$ to layer $J+1$ is as follows:
    
\begin{equation}
    \label{eq:18}
    X_{j+1}^t=MaxPool(ELU(Conv1d({[x_j^t]}_{AB})))
\end{equation}

    The operation of distillation here is slightly different from DeiT. It mainly uses 1D convolution and maximum pooling to build dimension and reduce memory usage before sending the output of the previous layer to the MHA module.
    
    \item Generative Decoder\hfill
    
    The structure of the generative decoder deviates from the standard decoder of Vanilla Transformer. Rather than following the conventional step-by-step approach, the generative decoder directly produces multi-step predictions in a single step. To achieve this, the output of the generative decoder is connected to a fully connected layer. The size of the output depends on whether univariate or multivariate prediction is needed.
\end{itemize}

The performance optimization of Informer has led researchers to apply it to anomaly detection tasks. Guo et al. \cite{RF40} pointed out three performance advantages of the Informer structure in the electrical line trip fault prediction task, that is,
\begin{itemize}
    \item The ProbSparse self-attention block has a good performance on sequential alignment;
    \item The self-attention distilling draws the main attention by halving the cascading layer to handle long input sequences efficiently;
    \item The generative style decoder predicts the long time-series sequence at one forward operation, which drastically improves the inference speed of the long-sequence output.
\end{itemize}

The GC method improves the training process compared to the Adam optimizer, and the projection gradient descent method with constrained loss function enables more efficient and stable training, further exploiting the performance advantages of Informer. TiSAT \cite{RF81} feeds a set of nominal historical observations into the Informer for training in an offline manner, and passes in the observations during testing until changes relative to the nominal baseline are detected. 

However, in extreme cases, Informer can also suffer from performance degradation. Therefore, researchers should carefully deliberate whether specific task requirements are applicable to the Informer model when performing transfer learning tasks.

\subsection{Anomaly detection based on Conformer \& Convolutional Transformer}

Conformer \cite{RF82} is a model proposed by Google for Automatic Speech Recognition (ASR) tasks. Since Vanilla Transformer is more effective in dealing with long sequence dependencies, while convolution is good at extracting local features, Conformer can improve both the long sequence and local features of the model by applying convolution to the Encoder layer of Transformer. The overall structure of Conformer is shown in Figure \ref{figure7}:

\begin{figure}[H]
    \centering
    \includegraphics[width=0.6\linewidth]{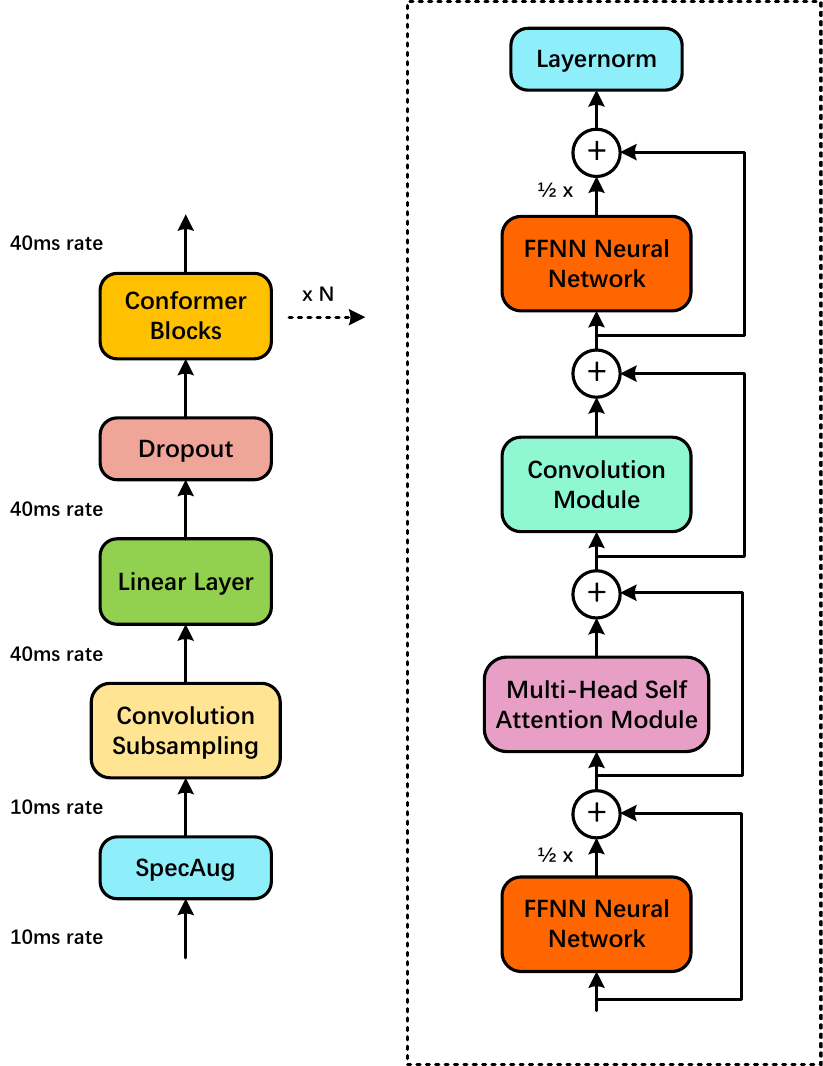}
    \caption{The overall structure of Conformer}
    \label{figure7}
\end{figure}

Conformer first carries out downsampling through the convolutional network, then passes through several Conformer blocks, each consisting of a $FFN$, MSA, and convolution module. For the convolution module, a gating mechanism (including point-by-point Convolution and linear gating unit) is first adopted, followed by a one-dimensional depth-separated convolution, and Batchnorm is used to help train deeper models. The convolution module can effectively extract local correlation. Finally, the feed-forward module consists of two layers of linear transformation and a nonlinear activation function, with additional use of the pre-norm residual unit and $Swish$ activation function. Therefore, assume that the $i^{th}$ input of the entire layer is $\tilde{x}_i$, then the output $y_i$ is obtained by the following calculation method:

\begin{equation}
    \label{eq:19}
    \tilde{x}_i=x_i+\dfrac{1}{2}FFN(x_i)
\end{equation}

\begin{equation}
    \label{eq:20}
    x'_i=\tilde{x}_i+MHSA(\tilde{x}_i)
\end{equation}

\begin{equation}
    \label{eq:21}
    x''_i=x'_i+Conv(x'_i)
\end{equation}

\begin{equation}
    \label{eq:22}
    y_i=Layernorm(x''_i+\dfrac{1}{2}FFN(x''_i))
\end{equation}

Conformer has the advantage of both local and global feature extraction capabilities. Stowell et al. \cite{RF83} used Conformer to extract sequence information from the entire audio input for DCASE2020 unsupervised anomaly sound detection. They also applied the anomaly score mechanism of the GMM method to improve anomaly detection accuracy and achieved excellent performance in this challenging task. Yella et al. \cite{RF84} adopted the Soft-Sensing Conformer for the wafer fault diagnosis classification tasks. They similarly concluded that Vanilla Transformer can effectively learn global location information for alleviating long-term dependency problems, but requires a large amount of data and suffers from quadratic complexity, while CNN can effectively learn local information. The process of their method is as follows: first, embedding the input, feeding the embedded data in the low dimensional space to the Conformer block. After receiving the output of multiple Conformer blocks, concatenating them and performing Global Averaging Pooling (GAP) operations. Finally, classifying the output through $Y=ActivationFunc (FFN(Dropout(z))$, and converting to the probability output through the $sigmoid$ function. The author additionally used the loss function $SuperLoss$ of the curriculum learning method to address the problems of noise and high data imbalance in traditional learning models. Their extensive experiments on various datasets from Seagate Technology's wafer manufacturing process demonstrated the superior performance of Conformer.

Some researchers did not directly use the Conformer model but combined Transformer with CNN. In this paper, these methods are collectively referred to as convolutional Transformers. Although different from Conformer in terms of model structure, both essentially use the ability of convolution to extract local features to compensate for the deficiency of Transformer. Li et al. \cite{RF64} used a convolution-based Transformer model for the main purpose of addressing the weakness of the Transformer's local perception. Their approach was to replace the linear projection unit in the Vanilla Transformer with depth-separable 1D convolution (DWConv1D), as shown in Figure \ref{figure8}:

\begin{figure}[H]
    \centering
    \includegraphics[width=0.57\linewidth]{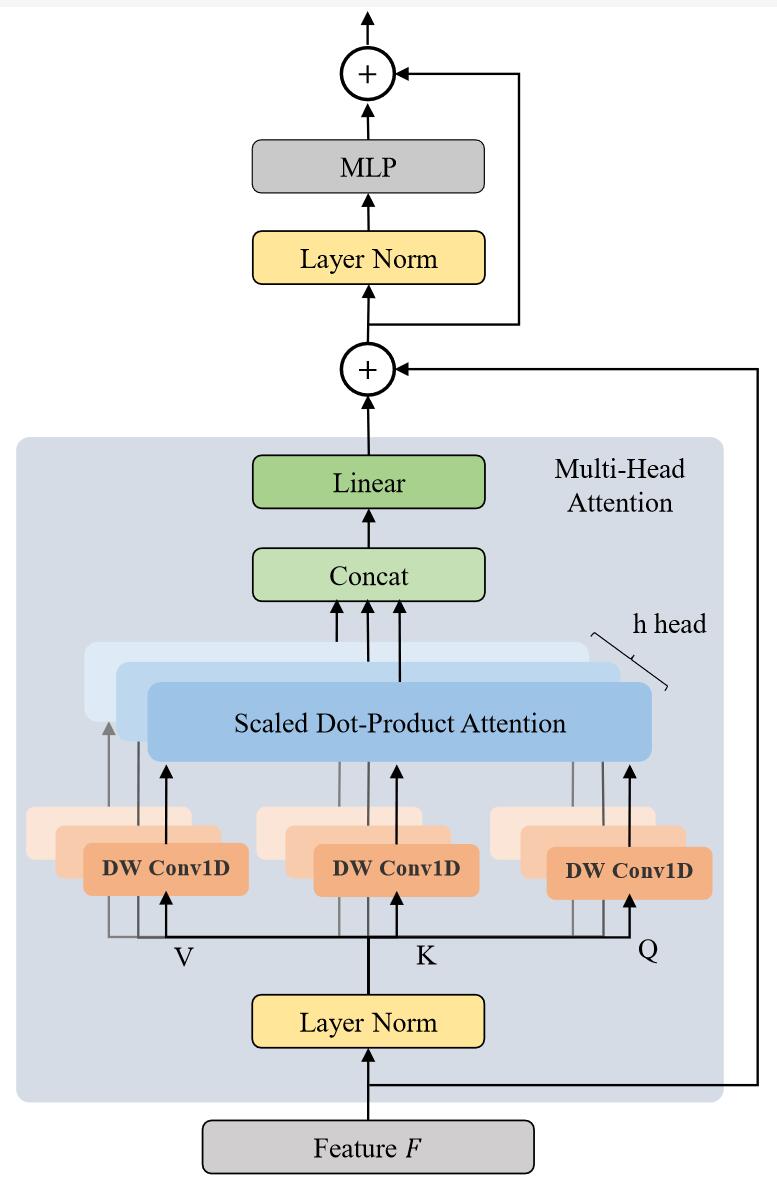}
    \caption{The structure of convolutional Transformer (NOTE: The figure is from the paper \cite{RF64})}
    \label{figure8}
\end{figure}

Multi-Sequence Learning essentially changes the single MIL video sequence segment to the average anomaly scores of consecutive segments. The whole model first extracts feature $F$ from videos containing $T$ segments through a backbone network, and then MSLNet takes $F$ as input, including a video classifier and a segment regressor. The video classifier is used to predict whether the video contains anomalies and contains two layers of Convolutional Transformer Encoder (CTE) and a linear layer. The segment regressor is used to predict the anomaly score of each video segment, which has the same structure as the video classifier. The final total loss is based on the hinge-based Multi-Sequence Learning ranking loss with classification loss. A score correction method is used to reduce the fluctuation of anomaly scores predicted by the segment regressor, i.e., making adjustments based on the prediction results of the video classifier, maintaining the anomaly scores if they are predicted to contain anomalies with high probability, and weakening the anomaly scores otherwise. In the training stage, a two-stage self-training mechanism is used. In the first stage, Multi-Sequence Learning is used to select the tag (as it is a weak-supervised learning). $Maxsi$ is used to choose the maximum average pseudo-label sequences so that the model has the preliminary ability to predict anomaly scores. In the second stage, MSLNet is used to select the sequence by prediction. Finally, the prediction scores are gradually refined by continuously halving the sequence length $K$ and repeating the above two stages. Tian et al. \cite{RF66} pointed out that the limitation of MIL is the inability to select rare anomalous segments in anomalous videos, so they used the contrastive snippet mining (CSM) algorithm to identify hard-to-recognize normal and anomalous clips and planned to further increase the online reasoning capability in the future.

Since 2022, convolutional Transformers have become even more of a research hotspot in multiple anomaly detection tasks. Researchers tend to incorporate Transformer as a crucial component in the entire algorithm pipeline (i.e., feature extraction, feature reconstruction, etc.). ADTR \cite{you2022adtr}, for instance, utilizes a frozen pre-trained CNN backbone to extract features and Transformer for feature reconstruction with an auxiliary learnable query embedding. They also proposed novel loss functions for image-level and pixel-level anomaly tasks, respectively. Jin et al. \cite{9854892} applied the convolutional Transformer to video anomaly detection tasks. They utilized a Transformer encoder to encode and extract the spatio-temporal features of the video and employed a convolutional decoder for anomaly prediction. Extracting spatio-temporal features is also a research focus of Sun et al \cite{sun2022transformer}. Similarly, Deshpande et al. \cite{deshpande2022anomaly} proposed utilizing a pre-trained videoswin Transformer model to extract better quality features and encode long and short-range dependencies in the temporal domain through attention layers. They performed anomaly detection using the Robust Temporal Feature Magnitude Learning (RTFM) model.

The SSMCTB method proposed by Madan et al. \cite{10273635} combines Transformer with CNN architecture. The enhanced Transformer Block can be simultaneously applied to multiple downstream tasks including image and video anomaly detection. ITran \cite{CAI2023106677} combines a multi-level feature extractor (ResNet-18 + multi-level jump connections) with Transformer encoder to put the anomaly detection work into the feature space and thus widen the generality gap between the reconstructed and original images. The TransCNN \cite{ULLAH2023106173} proposed by Ullah et al. utilizes a backbone CNN model to extract spatial features in the video and passes the features from the improved Transformer model to learn the long-term temporal relationships between various complex surveillance events. Kim et al. \cite{KIM2023105964} proposed a composite architecture for unsupervised time series anomaly detection tasks, combining a Transformer encoder with a decoder containing a 1D convolution layer. This architecture is capable of predicting input sequences taking into account both global trends and local variables.

For Conformer and convolutional Transformer models, the key drawback is that the introduction of convolutional operations further increases the complexity of the algorithm. Therefore, optimizing performance and improving efficiency for real-time online detection deployment is the pain point problem that these approaches should address.

\subsection{Anomaly detection based on Performer}

Performer \cite{RF85} is also designed to improve the performance and running efficiency of the Transformer. But unlike Informer, it does not rely on the sparsity feature of attention or low-rank matrices but uses Fast Attention Via positive Orthogonal Random features approach (FAVOR+) to reduce the computational complexity to a linear level.

Performer first modifies the attention calculation method in Transformer and adopts another equivalent expression, as shown in the following equation:

\begin{equation}
    \label{eq:23}
    Att(Q,K,V)=D^{-1}AV
\end{equation}

Where $A=exp(QK^T/\sqrt{d})$, $D=diag(A1_L)$. $diag$ is the operation of converting to diagonal matrix, $1_L$ is a vector with length $L$ of all 1s. Performer then introduces the kernel technique to approximate the attention matrix $A$. For any vector $q_i$ and $k_j$ in matrix $Q$ and matrix $K$, the kernel method is computed as:

\begin{equation}
    \label{eq:24}
    K(x,y)=E[\phi{(x)}^T\phi{(y)}]
\end{equation}

Where $\phi$ is a mapping from $D$ to $R$ dimensions. Therefore the key problem is finding a mapping of the function $\phi$ to reduce the computational complexity. In Performer, the function mapping $\phi$ is further defined as follows:

\begin{equation}
    \label{eq:25}
    \phi(x)=\dfrac{h(x)}{\sqrt{m}}(f_1(\omega_1^Tx),...,f_1(\omega_m^Tx),...,f_l(\omega_1^Tx),...,f_l(\omega_m^Tx))
\end{equation}

$f_1,...,f_l$ are different mapping functions, $\omega_1,...,\omega_m$ are sampled from the same distribution $D$. Performer uses the $softmax$ core for fitting

\begin{equation}
    \label{eq:26}
    SM(x,y)=exp(\dfrac{{||x||}^2}{2})K_{gauss}(x,y)exp(\dfrac{{||y||}^2}{2})
\end{equation}

To ensure that all weights are positive, the authors used an unbiased approximation. Performer further considers multiple sets of orthogonal $W$ parameters to further reduce the variance and obtain a better acceleration ratio. VQGAN \cite{RF86} compresses 3D information into discrete potential representations, then finds the nearest neighbor in codebooks composed of $K$ N-dimensional vectors by $L2$ norm, and replaces the representation with the nearest neighbor’s codebook index $K$ to quantify the representation. The decoder reconstructs the input from the quantized latent space, and attempts to distinguish the real image from the reconstructed image using a discriminator $D$. Performer learns the probability density of these representations, flattens out the discrete representations of $3D$ into $1D$ sequences, then estimates the image likelihood in terms of conditional probability, and generates a spatial likelihood graph $(1D \rightarrow 3D)$ by reshaping each image and up-sampling it. Here Performer can address the quadratic memory dependence of Transformer's attention mechanism on sequence length and the difficulty of training on large-scale sequences. By using a linearized approximation of the attention matrix, Performer can allow training on longer sequences. Experimental results show that Performer can be used simultaneously with the segmentation network with uncertainty. Performer can filter high OOD images, and the segmentation network can provide meaningful uncertainty estimates for images with only slight OOD. 

Pinaya et al. \cite{RF5} used an autoregressive Performer and VQ-VAE for brain anomaly detection and segmentation. Instead of directly applying Performer to learn the distribution over individual pixels, they compressed the input data into spatially small quantized potential projections through VQ-VAE, and subsequently fed into Performer for feature learning. They further extended the details about the experiments in their subsequent work \cite{PINAYA2022102475}.

\subsection{Anomaly detection based on Set Transformer}

Set Transformer \cite{RF50} is a Transformer model for processing set data with unrelated element order. (instead of Vanilla Transformer’s self-attention for sequential data) Set Transformer uses self-attention to process each element in the dataset, which forms a Transformer-like structure for modeling set-type data and can capture pairs or more complex interactions between elements. Given two matrices $X, Y\in R^{n\times d}$ for representing two sets with $N$ elements, each of which is a d-dimensional vector, then

\begin{equation}
    \label{eq:27}
    MAB(X,Y)=LayerNorm(H+rFF(H))
\end{equation}

Where $H=LayerNorm(X+Multi-head(X,Y,Y;\omega))$.
Here, $MAB$ is Multi-head Attention Block. $rFF$ is a layer that operates on a row, applying the same operation to each element. Set Transformer proposes $ISAB$, which is the optimized version of $MAB$, by introducing the matrix $I\in R^{m\times d}$, we can get

\begin{equation}
    \label{eq:28}
    ISAB_m(X)=MAB(X,H)\in R^{n \times d}
\end{equation}

Where $H=MAB(I,X)\in R^{m \times d}$.

The process is essentially a projection of $X$ into a lower dimensional space $H$ and then performs a reconstruction output of the higher-dimensional space. For the Decoder of Set Transformer, the equation is as follows:

\begin{equation}
    \label{eq:29}
    H=SAB(PMA_k(Z))
\end{equation}

Where $PMA$ is Pooling by MHA and $S\in R^{(k\times d)}$, which is to aggregate multiple features. After pooling, the relationship between $K$ outputs can be modeled through $SAB$. Therefore, the entire Encoder and Decoder can be represented by the following equation:

\begin{equation}
    \label{eq:30}
    Encoder=ISAB_m(ISAB_m(X))
\end{equation}

\begin{equation}
    \label{eq:31}
    Decoder(Z;\lambda)=rFF(SAB(PMA_k(Z)))
\end{equation}

The structural design of Set Transformer lacks universality and the model generalization ability is weak, so there are few related studies. Tajiri et al. \cite{RF49} used the Set Transformer’s permutation equivariant feature to effectively detect faults in ICT systems. Set Transformer can effectively handle records with missing values without interpolation, as well as variable-length records. 

\subsection{Anomaly detection based on BERT and its variants}

BERT \cite{RF15} is arguably one of the most important and widely used models since the introduction of Transformer. Based on BERT, many subdivision variants, such as RoBERTa \cite{RF87}, have been generated for different downstream tasks. BERT, which consists of stacked encoders, is a two-stage framework that starts with pre-training and then fine-tuning according to each specific task. In the pre-training stage, BERT requires a huge amount of data and consumes a lot of system resources. When the length of the training sentence is 512 Tokens, more than 40G of physical memory is often required to effectively support the parallel operation of BERT. Therefore, NLP open source communities such as HuggingFace \cite{RF88} have opened a series of pre-trained BERT models for the general research community to reduce the cost of using BERT models. BERT has two outputs: pooler output, corresponding output of $[CLS]$, and sequence output, corresponding to the last layer of the hidden output of all words in the sequence. Pooler output can be used for classification/regression tasks, and sequence output can be used for processing sequence tasks. BERT is only slightly different from the Vanilla Transformer Encoder in the input layer, which is modified to the following format:
\begin{equation}
    \label{eq:32}
    [CLS]+SENTENCEA(+SEP+SENTENCEB+[SEP])
\end{equation}

Where $[CLS]$ represents the special token for the classification task, which is the pooler output, and $[SEP]$ is the separator. $SENTENCEA$ and $SENTENCEB$ are both the input text of the model, where $SENTENCEB$ can be empty. BERT does not use the $\sin$ and $\cos$ function encoding method of Vanilla Transformer but obtains the position information through a method similar to word embedding. Since BERT may contain multiple sentence segment inputs, it is also necessary to add segment embedding.

Another innovation of BERT is the Mask Language Model (MLM) mechanism, which is the reason why BERT can be unconstrained by one-way language models. BERT predicts the original word at $[MASK]$ position by randomly replacing the token in each training sequence with a mask token at $15\%$ probability. BERT also performs the Next Sentence Prediction (NSP) task in the pre-training phase. Since MLM tasks tend to extract token-level representations, sentence-level representations cannot be obtained directly. To make the model capable of understanding the relationship between sentences, BERT uses NSP to predict whether two sentences are connected.

A large number of researchers have already applied BERT to anomaly detection tasks, most of which are for anomaly detection in log sequences \cite{chen2022bert}, due to the structural proximity of unstructured text-like log sequences to the NLP tasks handled by BERT. Based on HitAnomaly \cite{RF2}, Huang et al. further proposed HilBERT \cite{10070784}, a pre-trained log representation model with hierarchical bidirectional encoder Transformers. In addition, BERT itself can be used as a superior feature extractor for extracting anomalous features from logs. However, related researches also lead to some conflicting views, such as whether a pre-training task should be performed when using BERT for anomaly detection. These questions are explored in further detail in Section \ref{7.1}. 

You et al. \cite{RF89} used BERT to extract corresponding features from unstructured text data and adopted MLP for anomaly detection. They used a projected gradient descent method to enhance recognition by jointly minimizing the amount of normal class data and updating centers of the closed hypersphere, and further employing a smaller proportion of anomaly-labeled data. They considered the model’s anomaly detection capability by two application scenarios: detecting irrelevant sentences in the topic and detecting mislabeled text data that do not belong to any known class type. LAnoBERT \cite{RF51} focuses on the MLM mechanism. If the log sequence is normal, the prediction probability of the masked words is high, because it conforms to the log format learned by the model during the training phase. If the model has a lower prediction probability and a higher error (a larger distance metric between the predicted word and the originally masked word), the log is considered an anomaly log. However, LAnoBERT also reveals the shortcomings of the BERT model. Original BERT requires a lot of time to learn large-scale log data, and the training time should be shortened by effectively selecting the log data required for training and constructing a more lightweight BERT model to cope with the real-time anomaly detection task. In addition, it would cost too much performance to construct a BERT model for each dataset for anomaly detection. Therefore, a model that can accurately understand the nature of log data during the pre-training stage should be used, and fine-tuning operations should be carried out according to different log datasets, to improve the anomaly detection performance significantly. Le et al. \cite{RF34}, on the other hand, applied BERT to extract the raw log message semantics and converted them into semantic vectors. They argued that existing log parsing algorithms, such as Drain \cite{RF90}, can invalidate the entire anomaly detection approach due to an initial parsing error. Existing log parsing algorithms can not handle OOV words in new logs well, thus losing semantic information when detecting anomalies. In addition, current log parsing methods may generate errors due to semantic misinterpretation. However, some researchers also put forward a different viewpoint. This article will further discuss whether to use log parsing in Section \ref{7.1}.

Ott et al. \cite{RF91} evaluated the performance of several different language models, including BERT, GPT-2, and XL, on anomaly detection tasks for log data. They first preserved the semantics of log messages by pre-training these language models, mapped them to the embedding of log vectors, used Bi-LSTM neural networks to connect the embedding vectors through time, and applied the deviation between the detection system and the expected behavior to anomaly detection. To further improve the performance of anomaly detection, Hirakawa et al. \cite{RF92} used Optuna \cite{RF93}, a hyperparameter automatic optimization framework, and converted log messages into features (distributed expressions) rather than log keys. However, they concluded through experimental analysis that the current BERT model is still not very stable, and the performance is very sensitive to hyperparameters and learning rates. The size change of the dataset will further affect the performance of the model. Therefore, the BERT stable learning strategy should be further investigated in the future. TS-BERT \cite{RF94} uses MLM task to learn behavioral features of time series from a large amount of unlabeled data in the pre-training phase and uses the SR method to generate labels, which in turn eliminates the dependence of BERT on labels.  However, the author still found that the model performed better when partially labeled data was provided.

Anomaly Adapter \cite{RF95} uses RoBERTa to encode logs in byte form, uses Adapters to learn log structures and anomaly types, and additionally introduces transfer learning methods. The purpose of the Adapter is to fine-tune the structure of RoBERTa, so that the model does not have to be retrained for each data source, allowing the model to adapt to different anomaly datasets, similar to the fine-tune mechanism in BERT. Zhou et al. \cite{RF96} used the contrastive loss to fine-tune RoBERTa and improved the compactness of the representation. They also applied the method of Mahalanobis distance and contrastive learning to the RoBERTa model. Henrycks et al. \cite{RF97} comprehensively considered the performance of many BERT variants for the OOD detection task on NLP datasets. They measured the performance of BERT, BERT Large, RoBERTa, ALBERT \cite{RF98} and DistilBERT \cite{RF99} on various NLP tasks, including but not limited to emotion analysis, reading comprehension, etc. Both ALBERT and DistilBERT reduce memory consumption and improve the training speed of BERT.

The author's final evaluation results show that although the pre-trained BERT models are moderately robust to different OOD detection tasks, there is still room for further research and development. Podolskiy et al. took Hendrycks’ work one step further and proposed an OOD detection scheme using Mahalanobis distance in BERT \cite{RF100}. Through the fine-tuned BERT model, they could construct homogeneous representations of in-domain utterances and reveal geometric differences from out-of-domain utterances. which are easily captured by the Mahalanobis distance. While the fine-tuned BERT can allow a task-friendly reshaping of the embedding space. However, the authors argued that this method still suffers from its detection limitations. Firstly, it depends on the geometric features of the embedding space. For example, if the embedder is used as a classification model and overfitting at the same time, the geometric features of the embedding space may be destroyed. In addition, the biggest challenge of this approach is the semantically similar utterances, one of which is ID and the other is OOD. Xu et al. \cite{RF57} argued that previous models extract features from a certain layer in BERT, but this approach has performance limitations because the features of each layer in BERT model are different. If the features of all layers in BERT model are extracted directly, the efficiency is too low. If some feature dimension aggregation methods are used, such as Max pooling, information is sacrificed for efficiency. Therefore, the author used Mahalanobis distance to extract the potential representations of all layers, and automatically determine which layers’ features are important. Also, they used two methods, In-domain masked language modeling (IMLM) and Binary Classification with Auxiliary Datasets (BCAD) to fine-tune BERT within the domain. Experimental results show that pre-trained BERT can produce better feature representation. However, the simple aggregation of hierarchical structure is not very effective, while the fine-tuning approach further improves the overall performance of BERT.

\subsection{Anomaly detection based on other models}

In addition to the above methods, some researchers have proposed other variants of Transformer for different types of anomaly detection tasks. Although the methods themselves may have defects, these models have brought pioneering thinking to this research field. Network Transformer \cite{RF60} first uses a packet analyzer to group packets based on their source and destination addresses, uses Transformer to obtain an embedded representation, and then uses an aggregator to extract a series of hierarchical network features representing the traffic graph. By predicting $n$ future packets given several packet sequences, and training the anomaly detection algorithms at different abstraction levels (global, node, edge), it can facilitate anomaly detection and information extraction at different fine-grained levels, and enhance the learning ability of the system. Network Transformer can be applied to three different anomaly detection algorithms, namely LOF, OCSVM, and Autoencoder. Lee et al. \cite{RF101} investigated the performance of different neural network models including Transformer on the Virtual Network Function Chains anomaly detection task. The proposed model consists of a feature mapping layer, an encoder layer, a readout layer, and a classifier layer. At the encoder level, they compared several different models such as Transformer, Uni-RNN, Bi-RNN, etc. At the readout level, they tested several different functions including $max$, $mean$, and $self-attention$. They tested in two scenarios (i.e., web hosting service scenario and login authentication scenario), and considered the performance of joint training, i.e., training Transformer with several datasets at the same time. Experimental results fully demonstrated the robustness and superiority of Transformer. Skeleton-Transformer \cite{RF102} is a Transformer model combining MSA and Temporal Convolution Layer (TCL) for video anomaly detection tasks. MSA module captures the long-term dependencies of arbitrary paired pose components in spatial and temporal dimensions from different perspectives, while TCL focuses on local time information. Finally, the error between the predicted posture component and the corresponding expected value is used as the anomaly score. However, the disadvantage of this method is that it has very strict requirements on the datasets because the 2D skeleton trajectory extraction in the pre-processing stage must rely on high-quality video. Transformer for the Data Access Semantics (Trans-DAS) model in UCAD \cite{RF59} is improved based on Vanilla Transformer. The attention block of Trans-DAS uses a new masking mechanism, which connects an operation with its bidirectional operation context. The masking mechanism prevents inferring the semantics of the operation directly from the operation itself and allows Trans-DAS to capture the contextual intent of the operation based on its bidirectional context.

\subsection{Anomaly detection based on hybrid models}

Many methods combine Transformer model or its attention mechanism with other methods for anomaly detection tasks, which often complement the advantages of the Transformer, thus enhancing the overall performance of the model. These methods are collectively referred to as hybrid models in this paper. Mori et al. \cite{RF103} used the attention mechanism in Transformer for reference and applied it to RNN for anomaly noise detection. They used the model on multiple missing frames of a given log-scale MEL spectrum to compute the reconstruction errors of predicted and deleted frames as anomaly scores. TADDY \cite{RF104} combines the Vanilla Transformer with a discriminator in GAN to detect anomalies in dynamic graphs. TADDY mainly addresses two key difficulties in dynamic graph anomaly detection: the lack of informative encoding for unattributed nodes and the difficulty of learning to discriminate knowledge from coupled spatial-temporal dynamic graphs. TADDY designs a comprehensive node encoding that is inspired by the Position Encoding of Transformer model. It can simultaneously extract global spatial, local spatial, and temporal information, and integrate learnable mapping functions to help the framework automatically extract information encoding end-to-end. TADDY uses Transformer to learn spatial-temporal knowledge, adopts an edge-based substructure sampling method, takes cross-time contextual information as input, and then uses an attention mechanism to extract spatial-temporal coupling information. Xu et al. \cite{RF46} applied the minimax optimization strategy to Transformer. They alternately stacked anomalous attention blocks and feed-forward layers, which facilitated learning the underlying associations from deep multi-level features, and proposed anomalous attention modules with two branching structures.

Liu et al. \cite{RF105} combined heterogeneous information networks with Transformer for anomaly detection in smart contracts. They first extracted features to construct HIN and applied them to smart contracts. Then, they obtained the relationship matrix through the meta path learned by Transformer and used it as the input of CNN. Finally, they used node embedding for classification tasks. They adopted Doc2Vec to preprocess and convert the codes into vectors by using account features and code features as node attributes. Transformer is mainly used for encoding and decoding, converting the input data into output embedding. Experiments demonstrate the feasibility of Transformer for blockchain security applications. DCT-GAN \cite{RF48} uses the Transformer's encoder block to extract features in time series with additional GP and Lipschitz constraint methods. However, DCT-GAN does not fully exploit the performance of Transformer. It needs to first downscale into single-dimension data through PCA when processing MTS data, which can be very well solved by MHA in Transformer.

Intra \cite{RF106} is a self-supervised approach by performing the Inpainting task, i.e., covering certain regions of the image and then recovering them. Similar to other schemes used for image anomaly detection, Intra also separates the anomaly detection tasks into image-level fine-grained (anomaly detection) and pixel-level fine-grained (anomaly location). Intra uses both global and local Position Encoding because the global position information of image blocks does not need to be considered for texture class images but for some other types of images. Intra further introduces the MLP method for nonlinear dimensionality reduction because the authors found that the attention weights calculated by Vanilla Transformer were almost equivalent due to the similar features of different image blocks. This method is called Multi-head feature self-attention (MFSA). Although this scheme improves the accuracy of the model, it further increases the number of parameters in Transformer and prolongs the training time. GTA \cite{RF53} is sampled by the $Gumbel Softmax$ method to solve the problem that the BPTT algorithm could not be used due to the non-differentiable problem of discrete data sampling in classification distribution. They used Influence Propagation Convolution and Transformer's architecture to model time dependencies. Hierarchical extended convolution can effectively model sequences, set the multi-scale dilation size, and explore the temporal context modeling process with different sequence lengths and receptive fields. GTA re-modified Transformer's MHA and proposed a method called multi-branch attention. Multi-branch attention consists of two attention branches, a convolution branch for extracting information from a restricted neighborhood and another MHA branch for capturing long-range dependencies, and a multi-branch mixing strategy is proposed, which combines pairwise tokens with global learning attention. 

Wang et al. \cite{RF33} combined BERT and VAE to detect anomalies in log sequences. They used the Fast Gradient Method (FGM) to perturb the embedding layer of BERT model, generated the semantic features, and reduced the distance between the semantic features generated by log sequences before and after the perturbation of the embedding layer (a special method to enhance the robustness of the model). In addition, Contrastive Learning (making the semantic features generated by normal and anomaly log sequences have a larger gap and a longer distance in the semantic space) and VAE are used to extract statistical results to obtain statistical features. The statistical features and semantic features are combined to obtain enhanced features to train the model. Finally, MLP is used for anomaly detection tasks. However, this method configures different hyperparameters for different datasets to achieve optimal results, which has led to questions about the universality of the model. Li et al. \cite{RF107} proposed MFVT, a hybrid model combining FusionNet (essentially CNN structure) and ViT for anomalous traffic detection tasks. They fed the preprocessed data into FusionNet and ViT for training, iterated through Max Iteration, and finally obtained the trained model for evaluation using the Confusion Matrix. Their experiments illustrate that the combination of ViT and FusionNet can effectively address the challenges posed by unbalanced datasets, while also reducing the demand for sample data resources during training.

Zhang et al. \cite{RF108} applied Transformer's MHA to the Adversarial Learning OCC (ALOCC) model. MHA can effectively increase the difference between IC (Inlier class) and OC (Outlier class) samples, and solve the training instability problem of ALOCC. Gundawar et al. \cite{RF109} also noticed the excellent performance of MHA. They combined MHSA with LSTM and introduced Non-Parametric Dynamic Thresholding to classify certain points as anomaly thresholds, to realize anomaly detection of Spacecraft Telemetry. They also added improvements to the $sin$ and $cos$ Position Encoding functions of the Vanilla Transformer. By introducing the Time2Vec model \cite{RF110}, they expressed the $K+1$ input time series as Fourier series. Logsy \cite{RF111} is a method combining the self-attention mechanism with auxiliary data to improve log vector representation and perform unstructured log anomaly detection tasks. Logsy modifies the objective function through hyperspherical decision boundaries to achieve compact data representation and distance-based anomaly scoring. The author also used NLTK instead of existing log parsers to preprocess logs. Pereira et al. \cite{RF112} applied the Bi-LSTM model based on an attention mechanism to the task of anomaly detection of energy time series data. They adopted VAE and VASM mechanisms to introduce attention to the model. To enhance the robustness of the model, they also added noise to the input.

Yuan et al. \cite{RF113} combined ViT with U-Net and GAN, introducing TransAnomaly for video anomaly detection. TransAnomaly utilizes U-Net to encode spatial information and Transformer encoder to encode temporal information. It predicts future frames and calculates the difference with ground truth for anomaly detection. The Adapter of TransLog \cite{RF62} has a lightweight structure, consisting of a simple projection layer inserted in the middle of Transformer layer. When adjusting the model on downstream tasks, only the Adapter parameters are updated and the weight of the pre-trained model is frozen, which greatly reduces the trainable parameters of the model, but achieves considerable performance for transfer learning tasks. Zhang et al. \cite{RF114} combined Transformer with VAE for MTS anomaly detection tasks. They applied this hybrid model to a nonlinear state space, which can reduce the computational complexity, allow parallelization, and provide interpretable insights.

MT-RVAE \cite{RF54} improves the position encoding based on the Vanilla Transformer by using a global temporal encoding to add time series and period information to the data. This enhancement enables the model to capture long-term dependencies more effectively. Additionally, MT-RVAE introduces a multi-scale feature fusion algorithm for time series data. By integrating features from multiple time scales, the algorithm compensates for detailed information lost during the upsampling process, resulting in a more robust feature representation. Wang et al. proposed AnoDFDNet \cite{RF115}, a fusion model combining convolution and ViT for image anomaly detection of high-speed trains. They argued that the pre-trained models cannot detect "invisible" images, so AnoDFDNet does not employ the pre-training mechanism. AnoDFDNet can better distinguish the feature differences between anomaly images and normal images by detecting the divergence between two images taken at different times in the same area. The RT-SemiVAE model proposed by Chen et al. \cite{chen2023semisupervised} is designed for semi-supervised anomaly detection and localization in multivariate time series data. They achieved this by learning the long-term dependence of the data using a parallel multihead attention mechanism in the Transformer. Additionally, they utilized LSTM to capture short-term dependence. The introduction of parallel computing significantly reduces model training time. 

Tian et al. \cite{RF116} combined Transformer with MAE model \cite{RF117} and proposed a hybrid model called MemMC-MAE. The encoder part of MemMC-MAE incorporates a memory-enhanced self-attention operator, while the decoder utilizes a multi-level cross-attention mechanism. This combination allows for the correlation between normal patterns stored in the encoder memory and multiple normal patterns in the image to be effectively captured. By leveraging this correlation, MemMC-MAE can accurately reconstruct anomalies, resulting in high reconstruction errors. To address the issue of low anomaly image reconstruction error, the anomaly score in MemMC-MAE adopts the multi-scale structural similarity (MSSSIM) metric. Ma et al. \cite{RF118} first combined Transformer with federated learning techniques to perform privacy-preserving anomaly detection in cloud manufacturing. To avoid the conflict between Transformer and federated learning protocols, they designed a new collaborative learning protocol. Specifically, each edge device has a local encoder composed of Transformer structure, which extracts important anomaly data feature representations and uploads these encoded features with differential privacy noise to the cloud. The cloud contains a decoder with a MLP structure, which can distinguish between normal and anomalous features uploaded by edge devices. Further, the author used Gaussian noise to score the uploaded features instead of the raw data to detect anomalies, thus enhancing the privacy-protection ability of the framework and reducing communication overhead during training. However, the author indicated that the current model combining Transformer model and federated learning still has many shortcomings. For example, Transformer focuses too much on extracting more important features, which leads to a lower overall convergence speed of the model, and the distributed training protocol further limits the convergence speed. However, it is undeniable that the results strongly demonstrate the feasibility of combining Transformer with federated learning and provide a new research idea for the field of privacy protection anomaly detection.

Bi-Transformer Anomaly Detection method \cite{MA2023101949} first introduces the Binary Transformer architecture, which extracts temporal data correlation features from both the meta sequence and time-step sequence dimensions. It enhances detection performance through techniques such as generative adversarial training and MAML. This Binary Transformer architecture exhibits strong generalization capabilities and achieves high-precision detection on multiple time series datasets. Lakha et al. \cite{10020336} combined Transformer with GNN for anomaly detection in Cybersecurity events. They first used a GNN to produce representations of each event, which encode information about their neighboring events in an unsupervised manner. They then utilized complex features, such as command arguments that exhibit significant variation, which cannot be directly used as features in typical machine learning algorithms.

Kong et al. \cite{10083004} also combined Transformer with GAN for anomaly detection tasks in multivariate time series data. They modified the Transformer architecture and proposed the active distortion transformer (ADT) to capture temporal dependence and anomalous features by leveraging prior knowledge of overall associations. AnoFed proposed by Raza et al. \cite{RAZA2023106051}, combines the Transformer-based Autoencoder and VAE with Support Vector Data Description (SVDD) in a federated setting for anomaly detection in Electrocardiogram (ECG) images. They focused on the computational cost, making AnoFed efficient and lightweight, allowing it to be deployed on resource-constrained edge computing devices. Li et al. \cite{9956318} introduced the Memory-Token Transformer (MTT) to boost the reconstruction performance on normal frames. MTT takes the feature map and the memory module as input. A Transformer is then used to model the internal relationship between different memory tokens, projecting the final token to the size of the original feature map.

Qin et al. \cite{10021063} combined the Transformer model with signal decomposition. They provided a multi-view embedding method to capture temporal and correlational features of signals. They also designed a frequency attention module to extract periodic oscillation features. Finally, the signals were decomposed into seasonal, trend, and remainder components for further separate representation. The DADF \cite{10283467} model proposes, for the first time, the combination of Dual-Attention Transformer, Convolutional Block, and Discriminative flow. It integrates self-attention and memorial-attention mechanisms, incorporating sequential and normality associations. This integration enables the model to capture high-level global and local structural semantics of the inspected object. This approach innovatively gets rid of the limitations of previous reconstruction-based or distance-based anomaly detection methods, which tend to have high misclassification rates. By utilizing normalizing flow to learn a discriminative model over the joint distribution of the original features and two reconstructed features, it obtains the normality likelihood. This enables the simultaneous detection of local structural defects and global logical defects in anomaly detection tasks. RDAD \cite{xie2022rdad} combines the ideas of GAN and Transformer, composed of a reconstructive subnetwork and a discriminative subnetwork. The reconstructive subnetwork learns to reconstruct anomalous images without anomalies. In the discriminative subnetwork, the anomalous images and the reconstructed images obtained in the reconstructive subnetwork are concatenated as input. The reconstructed images assist the discriminative subnetwork in better inferring the anomaly of the input samples. RDAD further incorporates the squeeze-and-excitation block to enhance the sensitivity of relevant features by assigning attention to feature channels.

Yang et al. \cite{yang2022transformer} proposed an image anomaly detection method combining Transformer and GAN networks. They integrated an attention module into the depth-wise CNN-based encoder of the GAN to enhance the latent representation of input images. This approach addresses the limitation of the CNN encoder used in GAN-based methods for modeling the long-range information within image data, fusing global attention with local attention. Xiao et al. \cite{10005599} proposed the S2DWMTrans architecture, which deeply integrates convolutional operations with self-attention operations for anomaly detection in Hyperspectral images. S2DWMTrans consists of 3 main modules: DWMTrans, AWLF (Adaptive-Weighted Loss Function), and Postprocessing. DWMTrans is responsible for extracting spatial-spectral features from both global and local perspectives. AWLF helps control the training trend and further suppress anomalies. Postprocessing calculates the Mahalanobis distance for anomaly detection.

Liu et al. \cite{9926133} extensively redesigned the Transformer model and proposed a 2-stage based framework called "decouple and resolve" (DAR), consisting of the temporal proposal producer (TPP) and online anomaly localizer (OAL) modules. The TPP module aims to fully leverage hierarchical temporal relationships among snippets to generate precise pseudo-labels at the snippet level. Subsequently, using the fine-grained supervisory signals generated by TPP, the Transformer-based OAL module is trained to aggregate valuable cues from historical observations and anticipated future semantics to make predictions at the current time step. Both the TPP and OAL modules are jointly trained in a multi-task learning paradigm to share beneficial knowledge, enabling high-precision video anomaly localization tasks. For the same video anomaly detection task, Chen et al. \cite{chen2023mgfn} took a different approach and proposed the Magnitude-Contrastive Glance-and-Focus Network (MGFN). MGFN combines the $Q$, $K$, and $V$ matrices from the Vanilla Transformer with convolutional operations to form the Glance Block. They also introduced the Feature Amplification Mechanism and a Magnitude Contrastive Loss to enhance the discriminativeness of feature magnitudes for detecting anomalies.

In summary, hybrid models usually integrate Transformer with other auxiliary models (i.e., GAN, VAE, etc.) that already have excellent anomaly detection performance to compensate for the Transformer's disadvantages such as long training time and achieve better comprehensive performance.

\section{Application scenario of Transformer model in anomaly detection task} \label{5}

\begin{figure}[H]
    \centering
    \includegraphics[width=0.53\linewidth]{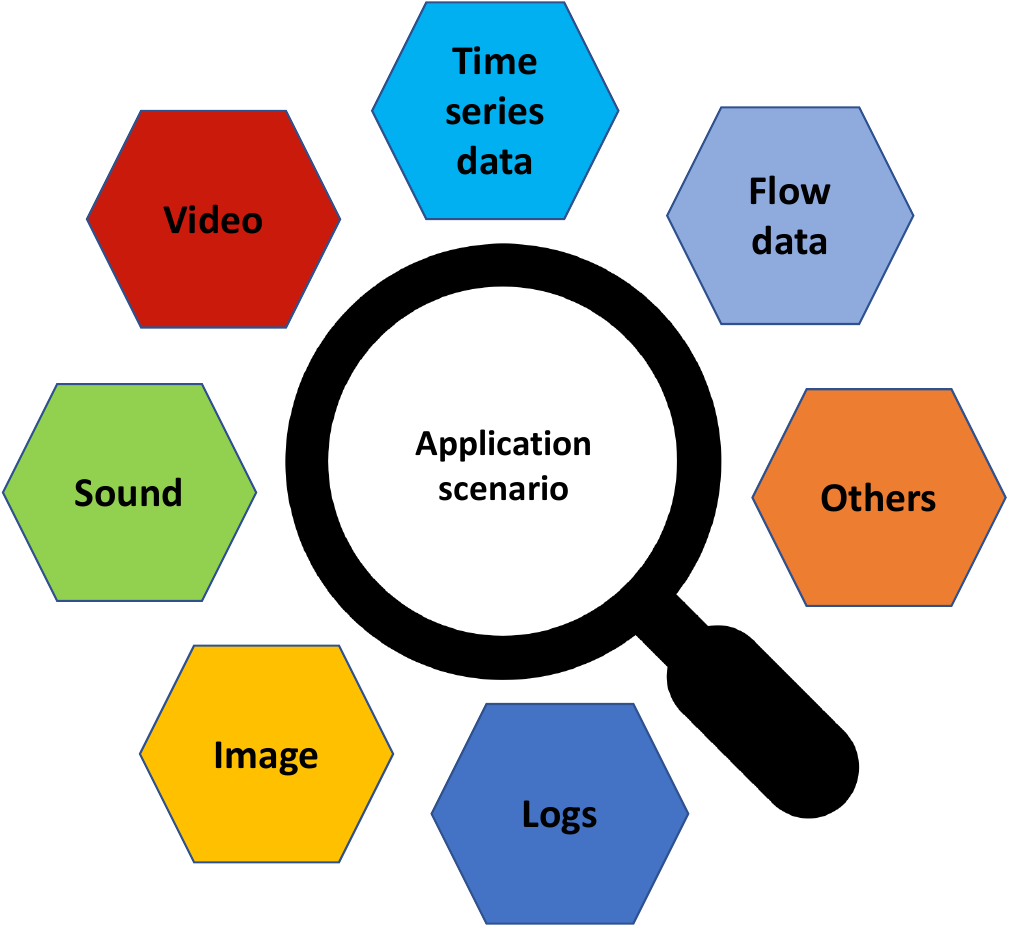}
    \caption{The application scenario of Transformer in anomaly detection task}
    \label{figure9}
\end{figure}

Unlike other anomaly detection reviews \cite{RF1} that classify the application scenarios into intrusion damage detection, medical, and public health anomaly detection, etc, this paper divides the application scenarios of anomaly detection into log, image, sound, video, time series data, flow data, and others according to data types, as shown in Figure \ref{figure9}.

The application scenarios of all research results collected in this paper are divided as shown in Table \ref{Table2}.

\begin{table}[h]\footnotesize
    \centering
    \caption{Application scenario division method of anomaly detection models}
    \begin{tabular*}{\linewidth}{cp{2.4cm}p{8.6cm}}
    \toprule
    Order & Application scenario \centering & References  \\
    \midrule
    1 & Logs and unstructured text anomaly detection & \cite{RF45} \cite{RF95} \cite{RF89} \cite{RF42} \cite{RF73} \cite{RF2} \cite{RF51} \cite{RF72} \cite{RF52} \cite{RF33} \cite{RF34} \cite{RF16} \cite{RF69} \cite{RF97} \cite{RF91} \cite{RF111} \cite{RF92} \cite{RF39} \cite{RF62} \cite{RF43} \cite{RF57} \cite{10138083} \cite{chen2022bert} \cite{10070784} \cite{10020336} \\
    \hline
    2 & Image anomaly detection & \cite{RF74} \cite{RF106} \cite{RF55} \cite{RF108} \cite{RF35} \cite{RF36} \cite{RF61} \cite{RF86} \cite{RF56} \cite{RF5} \cite{RF58} \cite{10054622} \cite{fan2023transformer} \cite{LIN2022104544} \cite{de2022masked} \cite{9858596} \cite{you2022adtr} \cite{CAI2023106677} \cite{PINAYA2022102475} \cite{RF115} \cite{RF116} \cite{10283467} \cite{xie2022rdad} \cite{yang2022transformer} \cite{10005599} \\
    \hline
    3 & Anomalous sound detection & \cite{RF103} \cite{RF83} (DCASE2020 challenge) \\
    \hline
    4 & Video anomaly detection & \cite{RF47} \cite{RF64} \cite{RF113} \cite{RF76} \cite{RF66} \cite{RF102} \cite{9774889} \cite{9956507} \cite{9854892} \cite{sun2022transformer} \cite{deshpande2022anomaly} \cite{ULLAH2023106173} \cite{9956318} \cite{9926133} \cite{chen2023mgfn} \\
    \hline
    5 & Time series anomaly detection & \cite{RF46} \cite{RF48} \cite{RF53} \cite{RF6} \cite{RF70} \cite{SIVAKUMAR2023100625} \cite{ding2023concept} \cite{RF81} \cite{RF18} \cite{ZENG2023244} \cite{RF94} \cite{RF112} \cite{RF114} \cite{RF54} \cite{KIM2023105964} \cite{chen2023semisupervised} \cite{MA2023101949} \cite{10083004} \cite{10021063} \\
    \hline
    6 & Flow data anomaly detection & \cite{RF3} (ICS system flow) \cite{RF68} \cite{RF107} \cite{RF60} \\
    \hline
    7 & Others & \cite{RF40} (Electrical Line Trip Fault) \cite{RF67} (IoT-networked environment) \cite{RF4} \cite{RF80} (ADS system) \cite{RF104} (Dynamic graphs) \cite{RF105} (Abnormal smart contract detection on Ethereum) \cite{RF49} (ICT system) \cite{RF30} (Network edge) \cite{RF32} (Network) \cite{RF109} (Spacecraft) \cite{RF75} (Traffic Scenario Infrastructures) \cite{RF38} (APT sequence) \cite{RF101} (Virtual Network Function Chains) \cite{RF84} (wafer fault-diagnostic classification) \cite{RF71} (Trajectory detection) \cite{RF60} (Database system) \cite{RF118} (Cloud manufacturing) \cite{RAZA2023106051} (ECG) \\ 
    \bottomrule
    \end{tabular*}
    \label{Table2}
\end{table}

(Note: The concept of anomaly detection here includes the concept category involved in the definition of anomaly detection discussed above, which belongs to anomaly detection in a broad sense.)

As can be seen from Table \ref{Table2}, since Transformer was initially applied to NLP tasks, a significant proportion of related research is carried out on unstructured text tasks such as logs, where Transformer has inherent performance advantages in this field. Transformer itself is not designed for audio modeling, so few relevant studies apply Transformer to audio anomaly detection tasks. These studies themselves do not directly use Transformer structure to model audio data but rather transform the audio into other forms of data representation \cite{RF103}. Due to the birth of ViT, a large number of studies have applied ViT to image anomaly detection tasks. Video anomaly detection is more difficult than image anomaly detection because video anomaly detection requires capturing not only the spatial feature information of each frame in the video but also the temporal feature information between different frames. Long and short-term dependencies are also a focal point of research in image/video anomaly detection. Therefore, Transformer often needs to be combined with other neural network models to better perform video anomaly detection tasks, and related research has only begun to show rapid development since 2022. However, the introduction of generalized frameworks based on hybrid models like SSMCTB \cite{10273635} have gradually revealed the capability for multimodal anomaly detection. (Which can be simultaneously applied to multiple downstream tasks such as image and video anomaly detection.) Time series data anomaly detection is also an important application scenario of Transformer, because Transformer can effectively model contextual information from a global perspective. Flow data anomaly detection task is similar to time series data anomaly detection task because flow can also be regarded as serialized information in essence. Transformer is also well qualified for the task of feature extraction and traffic modeling under its MHA. Numerous researchers have undertaken initial explorations in utilizing Transformer for anomaly detection tasks across various fields (e.g. the $7^{th}$ category of application scenarios). The endeavors have yielded promising outcomes, further highlighting the immense potential of Transformer in diverse anomaly detection tasks.

\section{Evaluation indexes and datasets of relevant research results} \label{6}
\subsection{Evaluation indexes}

At present, the mainstream evaluation indexes in the field of anomaly detection are $Precision$, $Recall$, $TPR$, $FPR$, $F1$ score, and $AUCROC$ (The $F1$-score and the area under the curve ($AUC$) of the receiver operating characteristics($ROC$) \cite{RF119}). Each index is defined as follows:

\begin{equation}
    \label{eq:33}
    precision=\dfrac{TP}{TP+FP}
\end{equation}

\begin{equation}
    \label{eq:34}
    recall=\dfrac{TP}{TP+FN}
\end{equation}

\begin{equation}
    \label{eq:35}
    F1=\dfrac{2\times precision \times recall}{precision+recall}
\end{equation}

\begin{equation}
    \label{eq:36}
    TPR=\dfrac{TP}{TP+FN}
\end{equation}

\begin{equation}
    \label{eq:37}
    FPR=\dfrac{FP}{FP+TN}
\end{equation}

\begin{equation}
    \label{eq:38}
    ACC=\dfrac{TP+TN}{TP+FP+FN+TN}
\end{equation}

\begin{equation}
    \label{eq:39}
    AUC=\dfrac{S_0-M \times (N+1)/2}{M \times N}
\end{equation}

Where $S_0=\sum r_i$. $TP$, $FN$, $FP$, and $TN$ are the number of True Positive, False Negative, False Positive, and True Negative respectively. $ROC$ is a probability curve, and $AUC$ value represents the degree and measure of separability. $M, N$ denotes the number of positive and negative examples respectively, and $r_i$ represents the ranking of the $i^{th}$ positive example in the ranking score list \cite{RF120}. If the $AUC$ value of a model is close to 1, it indicates that the model has excellent separability measurement performance. Table \ref{Table3} shows the evaluation indexes used by part of Transformer based anomaly detection models:

\begin{table}[h]\footnotesize
    \centering
    \caption{Evaluation indexes used by different researches}
    \begin{tabular}{@{}ccc@{}}
    \toprule
    No. & Authors & Index \\
    \midrule
    1 & kozik \cite{RF67} & Pre,Rec,F1 \\
    2 & Zhang \cite{RF3} & Acc,Pre,Rec,F1,ROC \\
    3 & Han \cite{RF4} & Pre,Rec,F1,AUC \\
    4 & Wittkopp \cite{RF45} & Pre,Rec,F1 \\
    5 &	Mori \cite{RF103} & AUC \\
    6 &	Liu \cite{RF104} & AUC,ROC \\
    7 &	Xu \cite{RF46} & Pre,Rec,F1,AUC,ROC \\
    8 &	Unal \cite{RF95} & Pre,Rec,F1 \\
    9 &	Feng \cite{RF47} & AUC,ROC \\
    10 & Li \cite{RF48} & ROC,F1 \\
    11 & Guo \cite{RF73} & Pre,Rec,F1 \\
    12 & Stowell \cite{RF83} & AUC,pAUC(part AUC) \\
    13 & Tajiri \cite{RF49} & ROC \\
    14 & Xu \cite{RF32} & Pre,Rec,F1 \\
    15 & Pirnay \cite{RF106} & AUC,ROC \\
    16 & Lee \cite{RF51} & F1,ROC \\
    17 & Chen \cite{RF53} & Pre,Rec,F1 \\
    18 & Wibisono \cite{RF72} & Pre,Rec,F1 \\
    19 & Zhang \cite{RF52} & Pre,Rec,F1 \\
    20 & Wang \cite{RF33} & Pre,Rec,F1 \\
    21 & Le \cite{RF34} &  Pre,Rec,F1 \\
    22 & Li \cite{RF107} & Pre,Rec,F1,ACC \\
    23 & Wrust \cite{RF75} & AUC \\
    24 & Koner \cite{RF36} & ROC \\
    25 & Yu \cite{RF38} & ACC,ROC,TPR \\
    26 & Nedelkoski \cite{RF111} & ACC,Pre,Rec,F1 \\
    27 & Yella \cite{RF84} & ROC \\
    28 & Hirakawa \cite{RF92} & ROC,F1 \\
    29 & Meng \cite{RF6} & Pre,Rec \\
    30 & Lee \cite{RF101} & F1 \\
    31 & Yuan \cite{RF113} & AUC \\
    32 & Guo \cite{RF62} & Pre,Rec,F1 \\
    33 & Zhao \cite{RF43} & Pre,Rec,F1,ACC,FPR \\
    34 & Schneider \cite{RF56} & AUC,ROC \\
    35 & Zhang \cite{RF114} & Pre,Rec,F1 \\
    \bottomrule
    \end{tabular}
    \label{Table3}
\end{table}

However, some researchers believe that the above mainstream evaluation indexes have drawbacks and deficiencies, and cannot effectively measure the anomaly detection performance of the model. Lobo et al. \cite{RF119} pointed out many problems in $ROC/AUC$ indexes. Therefore, some authors put forward improvement schemes based on the above evaluation indexes, or use variants of them. You et al. \cite{RF89} used the Mean Average Precision evaluation index ($MAP$), as well as $Recall@k$ and $AUC$. $MAP$ is defined as:

\begin{equation}
    \label{eq:40}
    MAP=\dfrac{1}{m}\sum\nolimits_{i=1}^m\dfrac{|anomalies\quad above\quad or\quad at\quad a_i|}{P(a_i)}
\end{equation}

Where $m$ is the total number of anomalies, $a_i$ represents the $i^{th}$ anomaly, which is the descending order of Euclidean distance to the center point, and $P(a_i)$ is the position of the $i^{th}$ anomaly in all test examples.

$Recall@k$ is defined as:

\begin{equation}
    \label{eq:41}
    Recall@k=\dfrac{|anomalies\quad above\quad or\quad at\quad k\%n'|}{m}
\end{equation}

It is used to capture the percentage of anomalies retrieved at $k\%$ of the test data sample. Where $n'$ is the total number of test samples. Intuitively, $Recall@k$ can be understood as the recall value of the data volume at $k\%$. $MAP$ and $Recall@k$ are improvements on precision and recall evaluation indexes to consider the comprehensive performance of the model under different data volumes. Liu et al. \cite{RF105} used $F1$, $Micro-F1$, $Macro-F1$, $NMI$ and $ARI$ evaluation indexes. $Micro-F1$ and $Macro-F1$ are both evaluation indexes used to measure multi-class classification tasks. Assuming that the total number of current categories is 3, $Micro-F1$ is calculated as follows:\\

First, we calculate the total 
\begin{equation}
    \label{eq:42}
    Recall_m=\dfrac{TP_1+TP_2+TP_3}{TP_1+TP_2+TP_3+FN_1+FN_2+FN_3}
\end{equation}

and then calculate the total 
\begin{equation}
    \label{eq:43}
    Precision_m=\dfrac{TP_1+TP_2+TP_3}{TP_1+TP_2+TP_3+FP_1+FP_2+FP_3}
\end{equation}

finally, we get 
\begin{equation}
    \label{eq:44}
    micro-F1=\dfrac{2\times Recall_m\times Precision_m}{Recall_m+Precision_m}
\end{equation}

The calculation method of Macro-F1 is as follows:\\

First, we calculate 
\begin{equation}
    \label{eq:45}
    F1-score_i=\dfrac{2\times Recall_i\times Precision_i}{Recall_i+Precision_i}
\end{equation}
for each category, then we get 
\begin{equation}
    \label{eq:46}
    macro-F1=\dfrac{F1-score_1+F1-score_2+F1-score_3}{3}
\end{equation}

Both $NMI$ and $ARI$ are the evaluation indexes of unsupervised clustering methods. $NMI$ is calculated as follows:
\begin{equation}
    \label{eq:47}
    NMI(\Omega,C)=\dfrac{I(\Omega;C)}{(H(\Omega)+H(C))/2}
\end{equation}

Where $I$ represents mutual information and $H$ represents entropy. Adjusted Rand Index ($ARI$) is a kind of correction to the $RI$ index. $ARI$ is calculated according to the contingency table. The calculation equation is:

\begin{small}
\begin{equation}
    \label{eq:48}
    \overbrace{ARI}^{Adjusted\quad Index}=\dfrac{\overbrace{\sum\nolimits_{ij}\begin{pmatrix}n_{ij}\\2 \end{pmatrix}}^{Index}-\overbrace{[\sum\nolimits_i\begin{pmatrix}a_i\\2 \end{pmatrix}\sum\nolimits_j\begin{pmatrix}b_j\\2 \end{pmatrix}]/\begin{pmatrix}n\\2 \end{pmatrix}}^{Expected\quad Index}}{\underbrace{\dfrac{1}{2}[\sum\nolimits_i\begin{pmatrix}a_i\\2 \end{pmatrix}+\sum\nolimits_j\begin{pmatrix}b_j\\2 \end{pmatrix}]}_{Max\quad Index}-\underbrace{[\sum\nolimits_i\begin{pmatrix}a_i\\2 \end{pmatrix}\sum\nolimits_j\begin{pmatrix}b_j\\2 \end{pmatrix}]/\begin{pmatrix}n\\2 \end{pmatrix}}_{Expected\quad Index}}
\end{equation}
\end{small}

Zhou et al. \cite{RF96} additionally used the $FAR95$ evaluation index. The calculation equation of the $FAR$ index is as follows:

\begin{equation}
    \label{eq:49}
    FAR=\dfrac{FP}{FP+TN}
\end{equation}
$FAR95$ is the probability that a negative example (OOD) is misclassified as a positive example (ID) at a $TPR$ of $95\%$. In this case, a lower value indicates better performance. Manolache et al.\cite{RF42} additionally used the $AUPR$ evaluation index. The biggest difference between $AUPR$ and $AUC$ is that it is suitable for highly unbalanced datasets. GTF \cite{RF30} additionally uses the $IRLbl$ evaluation index to quantify the degree imbalance in the dataset and the Matthews Correlation Coefficient ($MCC$) evaluation index. The calculation method is as follows:

\begin{equation}
    \label{eq:50}
    N=TN+TP+FN+FP
\end{equation}

\begin{equation}
    \label{eq:51}
    S=\dfrac{TP+FN}{N}
\end{equation}

\begin{equation}
    \label{eq:52}
    P=\dfrac{TP+FP}{N}
\end{equation}

\begin{equation}
    \label{eq:53}
    MCC=\dfrac{TP/N-S\times P}{\sqrt{P\times S\times (1-S)\times (1-P)}}
\end{equation}

$MCC$ coefficients can be classified as a measure of binary quality. Wu et al. \cite{RF55} used $AUC$ evaluation index for image-level fine-grained anomalies and per-region-overlap ($PRO$) \cite{RF121} evaluation index for pixel-level fine-grained anomalies. The $PRO$ index is employed to construct a performance curve by varying thresholds and calculating the area under the curve as a comprehensive evaluation metric. It quantifies $PRO$ values at different thresholds, assessing the model's classification performance for each threshold by measuring the relative overlap between the binarized connected domain and the ground truth graph. Zhang et al. \cite{RF108} additionally adopted Visual Information Fidelity ($VIF$) \cite{RF122} and Equal Error Rate ($EER$). Podolskiy et al. \cite{RF100} adopted $ROC$, $AUPRooD$, $FPR@X$, etc. $AUPRooD$ is essentially not significantly different from the calculation method of $AUPR$, while $FPR@X$ is calculated according to the difference of $X$. if $X$ is $ID$, then $ID$ is considered positive. If $X$ is $OOD$, then $OOD$ is considered positive. Li et al. \cite{RF64} used $AUC$ and Average Precision ($AP$) evaluation indexes. $AP$ is the average value of $Precision$. Li et al. \cite{RF39} proposed a new evaluation index $Parsing Accuracy$. The calculation method is as follows:

\begin{equation}
    \label{eq:54}
    PA=\dfrac{count(correct\quad event\quad ID\quad group)}{count(all\quad event\quad ID\quad group)}
\end{equation}

It is used to measure the ratio of correctly parsed log messages to the total number of log messages.

Doshi et al. \cite{RF81} systematically specified the shortcomings of existing evaluation indexes and proposed a series of evaluation indexes for time series anomaly detection, as shown below:
	
$Sequence Detection Delay$:

given $\tau_i$ as the start time of anomaly event $i$, $T_i\geq \tau_i$ as the alarm time, the author defined the average detection delay as:

\begin{equation}
    \label{eq:55}
    ADD=\dfrac{1}{S}\sum_{i=1}^S(T_i-\tau_i)
\end{equation}

Where $S$ represents the number of anomalous events. The author argued that since most anomalies indicate a series of events, it is critical to detect anomaly events within a specific period. The author additionally defined the maximum tolerable delay $\delta_{max}$.

$Sequence Alarm Precision$:

The author defined the calculation method of alarm precision as:

\begin{equation}
    \label{eq:56}
    P=\dfrac{1}{\hat{S}}\sum_{j=1}^{\hat{S}}1_{\{=\}}T_j\in \bigcup[\tau_i,\tau_i+\delta_{max}]
\end{equation}

The author argued that their matrix focuses on detecting the real anomaly sequence compared with the instance-based accuracy measurement, therefore it only focuses on accurately detecting the occurrence of anomalous events. If an alarm is raised between the start of the alarm, it is considered a false alarm. Where $1_{\{.\}}$ represents an indicating function, $\hat{S}=|\{T_j\}|$ indicates the number of all alarms, $|.|$ denotes the cardinality of the collection.

$Sequence Precision Delay$:

\begin{equation}
    \label{eq:57}
    SPD=\int_0^1P(\alpha)d\alpha
\end{equation}

$SPD$ combines sequence-based average detection delay with sequence alarm accuracy to achieve a metric for easy comparison of event sequence algorithms. $SPD$ statistically quantifies the accuracy and normalized ADD ($NADD$) of the area under the curves, which is similar to the $AUC$ evaluation index. To map $ADD$ to the interval $[0,1]$, the author performed a normalization operation. $NADD=ADD/\delta_{max}$. $\alpha$ denotes $NADD$, and $P$ indicates $Precision$.

Similar to Doshi et al., Dang et al. \cite{RF94} argued that time delay should be considered in the evaluation index of time series anomaly detection task. Therefore, they added additional time delay measurements based on $Pre$, $Rec$, and $F1$ evaluation indexes to verify the validity of the detection.

Ma et al. \cite{MA2023101949} innovatively proposed the $PerformanceScore$ index using the numerical fitting method, to better and more intuitively consider the performance of a model in terms of performance and training efficiency. The calculation is as follows:

\begin{equation}
    \label{eq:58}
    PerformanceScore=\dfrac{e^{(AUC+F1-1) \times K}}{\log_{10}Time}
\end{equation}

The numerator of $PerformanceScore$'s calculation formula adopts the exponential form to amplify the performance impact of the 2 evaluation indexes $AUC$ and $F1$. The purpose of subtracting 1 is that the $AUC$ and $F1$ values are meaningless if they are lower than $0.5$ because even if pure probability prediction is used, the results of the corresponding $AUC$ and $F1$ values are 0.5. The denominator takes a logarithm to the running time to weaken the impact of time on $PerformanceScore$. $K$ in the numerator is an adjustable parameter, whose purpose is to balance the relationship between performance and efficiency, and can be set to different values according to different demand scenarios. For example, in a scenario with high-performance requirements, the value of parameter $K$ should be increased, while in a scenario with high-efficiency requirements, the value of parameter $K$ should be decreased.

Pinaya et al. \cite{RF5} used $ROC$, $AUCPRC$, $FPR95$, $FPR99$, and $FPR999$ evaluation indexes to measure the performance of near OOD tasks and far OOD tasks. Wang et al. \cite{RF54} used $FAR$, Missing Alarm Rate ($MAR$) evaluation indexes. Similar to the calculation method of evaluation indexes such as $TPR$, $FPR$, $FAR$, $MAR$ is calculated as follows:

\begin{equation}
    \label{eq:59}
    MAR=\dfrac{FN}{FN+TP}
\end{equation}

Mishra et al. \cite{RF44} used different evaluation indexes based on different datasets. They used $PRO$ on MVTec and BTAD datasets and $AUC$ on MNIST datasets. Wang et al. \cite{RF115} additionally used Overall Accuracy ($OA$) and Intersection over Union ($IoU$) evaluation indexes.
$IoU$ is originally used as an evaluation index for semantic segmentation tasks, and its calculation equation is as follows:

\begin{equation}
    \label{eq:60}
    IoU=\dfrac{target\bigwedge precision}{target\bigcup precision}
\end{equation}

However, the calculation method of $OA$ is not much different from general $ACC$, which simply measures the overall accuracy under the multiple categories' fine granularity. Ma et al. \cite{RF118} additionally used the $AUC-PR$ evaluation index based on the $AUC-ROC$ index. The calculation method is analogous to that of the $AUC-ROC$ index, except that the two-dimensional horizontal and vertical coordinates are replaced by $Recall$ and $Precision$, and calculating the area under the curve. $AUC-PR$ is more suitable for weighting $Precision$ and $Recall$, as well as selecting the appropriate threshold and focusing on the application scenarios of positive samples.

To sum up, the commonly used evaluation indexes for Transformer based anomaly detection research can be summarized as follows:
\begin{enumerate}
    \item $Pre$, $Rec$, $F1$, $ROC$, $AUC$, and other evaluation indexes can be regarded as the most widely used and generalized evaluation indexes in the field of anomaly detection. Researchers can use such evaluation indexes to compare performance with other research results quickly and intuitively;
    \item In different application scenarios, several evaluation indexes are specific to a certain application scenario and improved based on general evaluation indexes. For example, $FAR$ and $MAR$ are effective evaluation indexes to measure OOD detection performance. While the corresponding evaluation indexes in time series anomaly detection tasks should consider the problem of detection delay. This paper suggests that researchers can use such improved evaluation indexes to measure the performance of anomaly detection in a certain application scenario more accurately;
    \item Many scientific research results have mentioned the problem of model efficiency in the experimental part. In the context of big data, time efficiency has become a non-negligible factor in measuring the performance of anomaly detection models. However, few of the scientific research covered in this paper has separately proposed an evaluation index that can comprehensively consider both time efficiency and theoretical performance. Therefore, this paper argues that putting forward a comprehensive evaluation index considering both efficiency and performance should be the research focus of performance evaluation in this field \cite{MA2023101949}.
\end{enumerate}

\subsection{Datasets}
For different anomaly detection scenarios, different models use different datasets. In addition, for a small number of research results, the datasets are synthesized manually or collected from a private experimental platform, and such types of datasets are not considered in this paper. This article only examines the mainstream open-source datasets and some authorized datasets that can be publicly accessed. According to the division of application scenarios in Section \ref{5}, the corresponding datasets are organized and summarized in Table \ref{Table4}.

\begin{table}[h]\footnotesize
    \centering
    \caption{Datasets used in partial researches}
    \begin{tabular*}{\linewidth}{p{0.5cm}p{11.7cm}}
    \toprule
    Order \centering & References \& datasets \\
    \midrule
    1 \centering & \cite{RF45} (BGL, thunderbird, spirit1) \cite{RF95} (HDFS) \cite{RF89} (IMDB, wikitext-2) \cite{RF42} (20 Newsgroups, AG News) \cite{RF73} (HDFS) \cite{RF2} (HDFS, BGL, Openstack) \cite{RF51} (HDFS, BGL) \cite{RF72} (Openstack) \cite{RF52} \cite{RF33} (HDFS, BGL) \cite{RF34} (HDFS, BGL, thunderbird, spirit) \cite{RF16} (HDFS, BGL, thunderbird) \cite{RF69} (BGL, thunderbird, spirit) \cite{RF97} (SST2, IMDB, Yelp Review Dataset, Amazon Review Dataset, STS-B, MSRvid, ReCoRD, MNLI) \cite{RF91} (OpenStack) \cite{RF111} (BGL, thunderbird, spirit) \cite{RF92} (BGL) \cite{RF39} (BGL, HDFS, Android) \cite{RF62} (HDFS, BGL, Thunderbird) \cite{RF43} (HDFS, OpenStack) \cite{RF57} (Cross-corpus dataset SST, CLINIC150) \\
    \hline
    2 \centering & \cite{RF74} (MVTecAD, BTAD, CIFAR-10) \cite{RF106} (MVTecAD) \cite{RF55} (MVTecAD, MTD (Magnetic Tile Defects)) \cite{RF108} (UCSD, MNIST, COIL-100) \cite{RF35} (CIFAR-100, CIFAR-10) \cite{RF36} (CIFAR-100, CIFAR-10, ImageNet30) \cite{RF86} (CROMIS, KCH) \cite{RF56} (TIMo) \cite{RF5} (MedNIST) \cite{RF58} (MVTecAD, Retinal-OCT, Head-CT, Brain-MRI) \cite{RF115} (Diff, FB, OL) \cite{RF116} (HyperKvasir, Covid-19 Chest X-ray datasets) \\
    \hline
    3 \centering & \cite{RF103} (MIMII) \cite{RF83} (DCASE2020 challenge) \\
    \hline
    4 \centering & \cite{RF47} (UCSD Ped2, CUHK Avenue Shanghai Tech)  \cite{RF64} (Shanghai Tech, UCF-Crime, XD-Violence) \cite{RF113} (CUHK, UCSD) \cite{RF66} (Hyper Kvasir, LDPolypVideo) \cite{RF102} (HR-Shanghai Tech) \\
    \hline
    5 \centering & \cite{RF46} (SMD, PSM, MSL, SMAP, SwaT, NeurIPS-TS) \cite{RF48} (NAB, SWaT, WADI) \cite{RF53} (SWaT, WADI, SMAP, MSL) \cite{RF6} (MSL) \cite{RF70} (SMAP, MSL) \cite{RF81} (SMD, PSM, MSL, SWaT) \cite{RF18} (NAB, UCR, MBA, SMAP, MSL, SwaT, WADI, SMD, MSDS) \cite{RF94} (KPI, Yahoo) \cite{RF114} (SMAP, MSL, SMD) \cite{RF54} (SKAB, NAB, SAT) \\
    \hline
    6 \centering & \cite{RF107} (IDS2017, IDS2012) \\
    \hline
    7 \centering & \cite{RF67} (Aposemat IoT-23) \cite{RF4} \cite{RF80} (Baidu Apollo, GTSRB, Tusimple) \cite{RF104} (UCI Messages, Digg, Emial-DNC, Bitcoin-Alpha, Bitcoin-OTC, AS-Topology) \cite{RF105} (http://etherscan.io) \cite{RF30} (KDD99, UNSW-NB15) \cite{RF32} (NAD) \cite{RF59} (HDFS, BGL, Thunderbird) \cite{RF118} (NSL-KDD, Spambase, Arrhythmia, Shuttle) \\ 
    \bottomrule
    \end{tabular*}
    \label{Table4}
\end{table}

It is easy to see from Table \ref{Table4} that there are very mature publicly available datasets in anomaly detection of log, image, video, and time series data, which are used by a wide range of researchers, such as log datasets BGL, Thunderbird, spirit, HDFS, OpenStack, etc., image datasets MNIST, MVTecAD, CIFAR, etc., video datasets CUHK, UCSD, etc., time series datasets SMD, MSL, SMAP, SWaT, WADI, NAB, etc. Using these kind of general datasets, researchers can quickly and intuitively compare the performance with other scientific research results. However, in other application scenarios, such as medical and ADS fields, researchers encounter diverse datasets that are challenging to uniformly quantify. This paper suggests that research scholars should try to make the datasets publicly available for easier access and performance comparison by other scholars. For example, the work by Jin et al. \cite{9854892} provides a high-quality test dataset for the task of unmanned aerial vehicles (UAV) video anomaly detection. In addition, due to the high dependence of Transformer models on the quantity and quality of data, relevant research also needs to consider the "data hungry" \cite{CAI2023106677} problem during model training, such as modifying the model structure or proposing new bias functions.

\section{Discussions} \label{7}
\subsection{Problems of existing research methods}\label{7.1}
\subsubsection{Pre-trained model VS Non-pre-trained model}
From the current research, researchers have disagreed on whether Transformer model needs to be pre-trained. Some researchers believe that Transformer itself has the defects of many parameters, slow training speed, and excessive resource consumption. However, most parameters of Transformer can be frozen through pre-training, and only a few parameters need to be fine-tuned according to the actual task during the formal training stage. As mentioned in the previous section, some researchers have designed their fine-tuning schemes, such as Adapter \cite{RF62}. However, some researchers have also started training the model with only initialized parameters for different anomaly detection tasks, and even fine-tuned the hyperparameters according to different test datasets, to achieve higher performance on different tasks.

In this paper, we argue that to solve this problem, it should depend on the specific task requirements. For some anomaly detection tasks that are easy to perform transfer learning and have similar features, such as log anomaly detection and unstructured text anomaly detection, etc., a pre-trained Transformer should be used. Through the pre-training stage, the model can learn in advance some common features of domain-related tasks, such as log templates. In the formal training stage, the model only needs to be fine-tuned according to the actual task requirements, thus greatly improving the training efficiency of the model. In addition, the pre-training stage also helps the model obtain more prior information and improve the model's decision-making performance. Pre-trained models can also perform transfer learning strategies to adapt to more anomaly detection tasks within the same domain. For example, Bozorgtabar et al \cite{bozorgtabar2023attention}. proposed a Masked Image Modeling (MIM) strategy, namely Attention-conditioned Patch Masking (APMask) for ViT-based pretraining by masking non-salient image patches that help the model to capture the local semantics while preserving the crucial structure associated with the foreground object. They further harnessed the self-attention map extracted from the Transformer to mask non-salient image patches without destroying the crucial structure associated with the foreground object. Subsequently, the pre-trained model is fine-tuned to detect and localize simulated anomalies generated under the guidance of the Transformer's self-attention map. However, pre-trained models must be supported by high-quality datasets to achieve significant results. For researchers, how to collect massive data suitable for model pre-training and how to choose the appropriate transfer learning strategy are the key issues they have to consider. The pre-training step should not be performed for those anomaly detection tasks with significant feature variations, or feature uniqueness and specificity, i.e., wafer fault anomaly detection, pedestrian trajectory anomaly detection, etc. In this case, if the model performs the pre-training phase, it will be interfered by the pre-training parameters during the formal decision-making task, and the model performance will be reduced due to the large difference between the data features at the pre-training stage and actual tasks. Google Brain team has also explored this issue on the ImageNet dataset \cite{RF123}.

\subsubsection{Log Parser VS Non-Log Parser}

In the context of log anomaly detection tasks, researchers have discussed the log parsing problem. Some researchers believe that log parsers can efficiently convert unstructured data into a unified structured format, which is convenient for processing by neural networks. Therefore, even if log parsers have the drawback of inaccurate parsing, existing log parsers should be used to parse the logs. Some researchers argue that if log parsers are used, some log parsing errors are caused by these parsers, and these errors cannot be recognized by the Transformer model itself, thus reducing the overall performance of the model. Table \ref{Table5} summarizes the views of relevant studies on whether to use log parsing or not.

\begin{sidewaystable}\footnotesize
    \centering
    \caption{Views of different researches on log parser}
    \begin{tabular*}{\textheight}{@{\extracolsep\fill}cccp{10cm}}
    \toprule
    Ref. & Whether used log parser & log parser name (if used) & Reasons \\
    \midrule
    \cite{RF51} & N & -- & First, for the models using a parser to preprocess log keys, a pre-defined template is required for converting the logs to a standardized template, and experts were needed to manually set the rules for refining log keys. The greatest problem of parser-based log anomaly detection is that certain crucial information can get lost during the standardization process. Furthermore, when a log parser is used, the performance of the log anomaly detection model becomes highly dependent on the compatibility of a log parser rather than the logic of the model itself. \\
    \hline
    \cite{RF33} & Y & Drain & The algorithm builds a parse tree based on the content of the log entry, and uses the information contained in each layer of the parse tree to determine the log template, thereby converting unstructured log entries into structured log templates.\\
    \hline
    \cite{RF34} & N & -- & Through our empirical study, we find that existing log-based anomaly detection approaches are significantly affected by log parsing errors that are introduced by OOV (out-of-vocabulary) words and semantic misunderstandings. The log parsing errors could cause the loss of important information for anomaly detection. \\
    \hline
    \cite{RF16} & Y & Drain & In order to represent log messages, following a typical pre-processing approach, we first extract log keys (string templates) from log messages via a log parser.\\ 
    \hline
    \cite{RF91} & Y & Drain & Log parsing provides a mapping function of the raw log messages into log templates, e.g. log instruction in the source code. In this work, we adopt Drain, due to its speed and efficiency. \\
    \hline
    \cite{RF111} & N & -- & However, the learning of the sequence of templates still has limitations in terms of generalization for previously unseen log messages. The methods tend to produce false predictions mostly owing to the imperfect log vector representations. \\
    \hline
    \cite{RF39} & Y & A method proposed by the paper. & SwissLog targets at those faults resulting in log sequence order changes and log time interval changes. To achieve that, an advanced log parser is introduced.\\
    \hline
    \cite{RF62} & Y & Drain & The purpose of log parsing is to convert unstructured log data into the structured event template by removing parameters and keeping keywords. \\ \hline
    \cite{RF43} & Y & Methods inspired by BERT & We design a log parsing method combining the log key and parameter values with semantic representation to alleviate the problem of insufficient information in event templates. \\
    \bottomrule
    \end{tabular*}
    \label{Table5}
\end{sidewaystable}

As can be seen from Table \ref{Table5}, researchers hold different views on whether to use a log parser and what type/method to use. Li et al. \cite{RF39} further classified log parsing methods into several categories, e.g., similarity-based clustering, frequency-based clustering, and heuristic methods based on search trees. In this paper, we argue that whether to use a log parser depends on the specific situation. Generally speaking, if a log parsing method such as Drain is used, the generality of the model can be enhanced. This is because a general log parsing method can convert log datasets from different sources and types into a uniform format that can be easily processed by neural network models, but parsing errors of general log parsers are inevitable. Self-proposed log preprocessing/parsing methods often achieve better performance than the general log parsing method on specific test datasets, but the generality of the model is not guaranteed. For self-designed log parsing methods, researchers need to consider the universality and generalization ability of the method, rather than just obtain excellent performance in several experimental environments mentioned in the paper.

\subsubsection{Reconstruction-based VS prediction-based, embedding-based method, etc.}

Several papers mentioned the classification of anomaly detection methods. Pirnay et al. \cite{RF106}, Wu et al. \cite{RF55} divided the methods of image anomaly detection into reconstruction-based and embedding-based methods. Chen et al. \cite{RF58} classified image anomaly detection into reconstruction-based methods and pre-training feature-based methods. Chen et al. \cite{RF53}, and Yuan et al. \cite{RF113} classified deep learning methods for MTS data anomaly detection and video anomaly detection into reconstruction-based methods and prediction-based methods, respectively. As seen from the above classification, the specific classification methods vary due to different anomaly detection tasks. But generally speaking, most researchers agree to classify anomaly detection methods into mainstream methods based on reconstruction, embedding, or prediction. There is no significant performance difference between these methods because the specific performance depends on the details of the model's internal implementation, parameter selection, training \& fine-tuning methods, etc. The classification of these methods is only different in a broad sense. However, these methods also have their limitations. For example, reconstruction-based methods tend to recover both normal and anomalous samples well, making it difficult to detect outliers. To address the issue of the identical shortcut in reconstruction-based methods, You et al. \cite{NEURIPS2022_1d774c11} proposed a unified model for multi-class anomaly detection. Firstly, they introduced a layer-wise query decoder to help model the multi-class distribution. Secondly, they employed the neighbor masked attention module to further avoid the information leak from the input feature to the reconstructed output feature. Thirdly, they proposed a feature jittering strategy that urges the model to recover the correct message even with noisy inputs.

\subsubsection{Distance measurement: Mahalanobis distance VS CE, $Cos$ distance and other loss functions}

Recently, many researchers began to apply Mahalanobis distance to the task of anomaly detection. They used Mahalanobis distance instead of traditional distance measurement methods such as CE and Euclidean distance and agreed that Mahalanobis distance can obtain more accurate distance measurement than traditional distance measurement methods, thus improving the accuracy of anomaly detection. Zhou et al. \cite{RF96} separately compared Mahalanobis distance with distance measures such as Maximum Softmax Probability ($MSP$), $Cos$, etc. in the experiment part, and the results showed that Mahalanobis distance has a performance advantage in $AUROC$, $FAR95$, and other evaluation indexes of 4 datasets. Koner et al. \cite{RF36} showed experimentally that the performance of Mahalanobis distance is similar to $cos$ distance, the former holds only a weak performance advantage. They attributed this to the smaller dependence of the higher norm between the two features of the Mahalanobis distance and the utilization of mean and covariance. Podolskiy et al. \cite{RF100} tested different combinations of models and distance functions through a large number of experiments, and finally concluded that the combination of fine-tuned Transformer and Mahalanobis distance could achieve the best performance. They believed that Mahalanobis distance can easily capture the geometric differences between the homogeneous representations of utterances inside and outside the domain. Xu et al. \cite{RF57} argued that Mahalanobis distance could effectively extract the features of all layers in the BERT model.

Combined with the above conclusions, this paper argues that there is still a certain performance advantage of Mahalanobis distance in Transformer-based OOD detection tasks, but it is not in the absolute lead. More research results are needed to support this conclusion in the future.

\subsection{Challenges}

\subsubsection{Challenges of anomaly detection tasks}

\begin{figure}[H]
    \centering
    \begin{minipage}[c]{0.35\linewidth}
    \centering
    \includegraphics[width=1.0\linewidth]{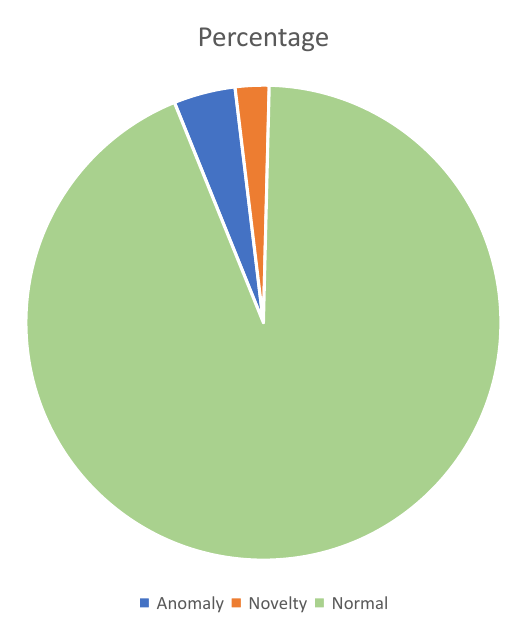}
    \label{fig10:a}
    \end{minipage}%
    \begin{minipage}[c]{0.35\textwidth}
    \centering
    \includegraphics[width=1.0\linewidth]{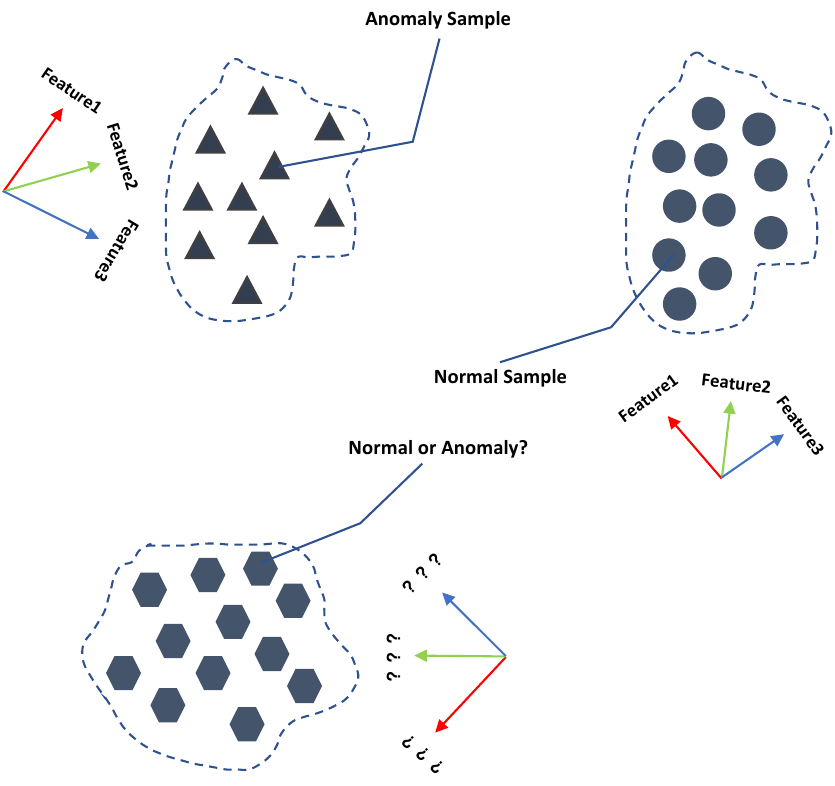}
    \label{fig10:b}
    \end{minipage}
    %\qquad
    
    \begin{minipage}[c]{0.35\textwidth}
    \centering
    \includegraphics[width=1.0\linewidth]{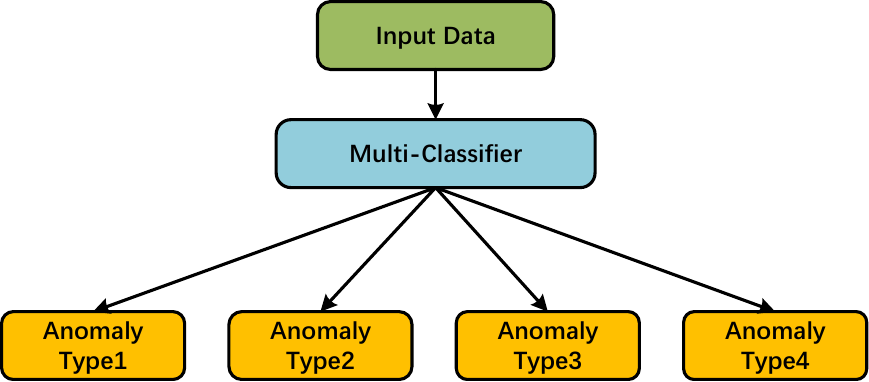}
    \label{fig10:c}
    \end{minipage}%
    \begin{minipage}[c]{0.35\textwidth}
    \centering
    \includegraphics[width=1.0\linewidth]{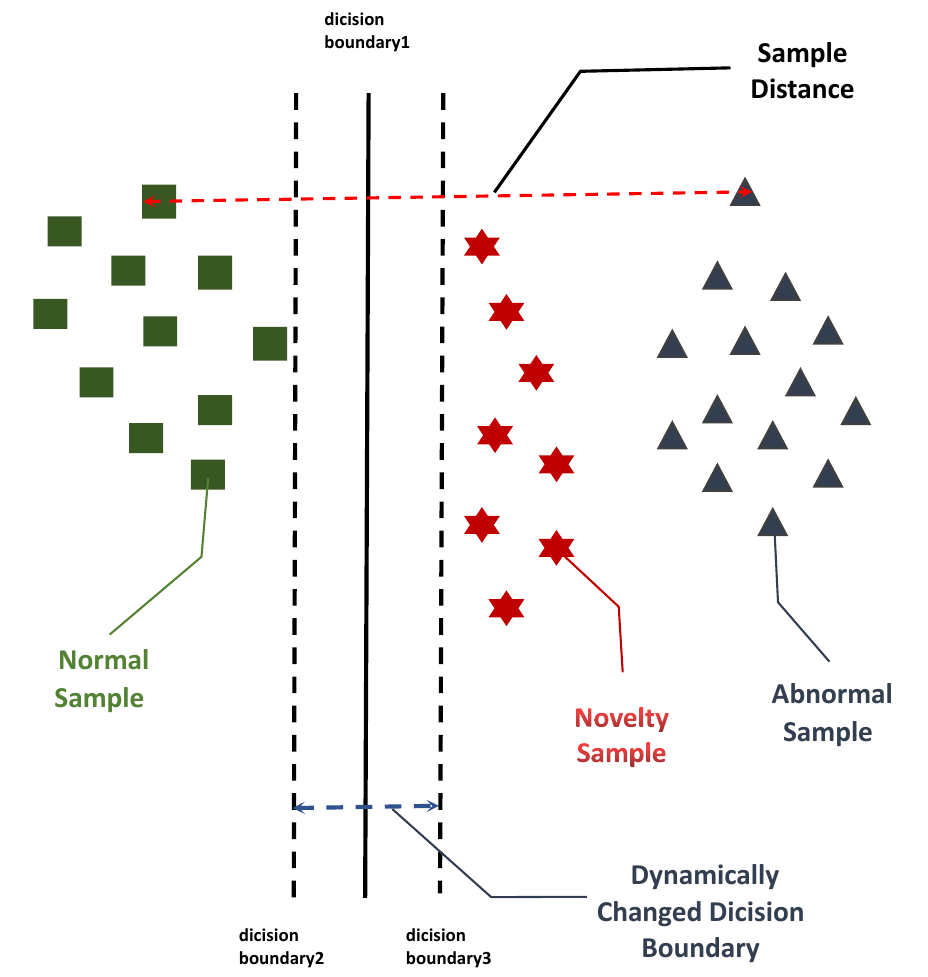}
    \label{fig10:d}
    \end{minipage}%
    \caption{4 main challenges faced by anomaly detection tasks}
    \label{figure10}
\end{figure}

Currently, various neural network models, including Transformer, are very prone to model distribution bias and model estimation errors, which is because the vast majority of samples in the task of anomaly detection are normal samples. This paper summarizes several main challenges faced by the current anomaly detection task, as shown in Figure \ref{figure10}.

\paragraph{Data imbalance.} \hfill

Most anomaly detection datasets contain a large number of normal samples and a very small number of abnormal samples, while there is also an imbalance in the internal distribution of most positive samples. Therefore, for the anomaly detection model, it is necessary to learn not only the appropriate representation from the training data but also the prior inductive deviation according to more reasonable constraints.

\paragraph{The interpretability of the model.} \hfill

At present, an important challenge for anomaly detection models is the problem of model interpretability. Since an important application of anomaly detection is to give the corresponding processing method based on the analysis of anomalies, it is not enough to give anomaly detection results for the corresponding model. The interpretability of a model affects its reliability and also facilitates more related tasks of anomaly detection, such as anomaly prediction, anomaly diagnosis, anomaly location, etc. Anomaly detection models should have a clear hierarchical, cascading structure to quantify and infer anomalies in an interpretable manner.

\paragraph{Multi-class classification and overfitting problem.} \hfill

Currently, most anomaly detection models are designed for binary classification problems. However, with the development of modern industrial systems and the increasing complexity of system architectures, there are higher requirements for anomaly detection tasks: a system should not only be able to detect anomalies in data sequence, but also give specific information about anomaly type. For a model, this entails the ability to effectively tackle multi-class classification tasks, learn distinctive features of various anomaly types from limited datasets, generalize this knowledge to accurately classify diverse anomalies, and offer appropriate indications during testing. This task is not as easy as adding the $Softmax$ multi-classification layer in the model output part, because the model needs to learn different anomaly features and identify different types of anomalies in various datasets and application scenarios, which requires strong learning capability. Therefore, only a very few studies, such as GTF \cite{RF30} and OODformer \cite{RF36} have conducted experiments on multi-class classification tasks. Multi-class classification tasks require models that can accurately identify several different types of anomaly features and make correct decisions. Therefore, current multi-class classification anomaly detection models are all learned through supervised training. These models are trained under scenario-specific, data-limited datasets with multi-class classification labels, with weak generalization ability and unstable parameter selection. The data imbalance problem further limits the development of such models. Consequently, the multi-class classification problem is one of the severe challenges faced by Transformer models. In Section \ref{7.3.4}, we put forward further development directions for multi-class classification problems.

For Transformer, the goal is to produce a compact representation that can accurately represent normal samples while difficult to represent anomalous samples. However, if we restrict the latent space too much, it will lead to overfitting problems and is not conducive to the generalization of the model. Therefore, correct representation learning is crucial, especially in the context of data imbalance, where decision boundaries are difficult to determine. In this case, the choice of representation will affect the shape of normal distribution, and then affect the performance of anomaly diagnosis. Thus, selecting appropriate representation methods and solving the overfitting problem is one of the important challenges faced by Transformer. 
 
\paragraph{Decision boundary ambiguity problem.} \hfill

Model decision boundary should be equivalent to the ideal distribution boundary under ideal circumstances. However, the deviation and distortion of the model will lead to ambiguity between abnormal samples, normal samples, and new samples. (Note that the boundary ambiguity here does not mean that the model decision boundary is uncertain, but that too close sample spatial distances will make it difficult for the model to determine a clear decision boundary.) Model needs to widen the distance between normal and anomalous samples during learning representation to adaptively use appropriate decision boundaries for different anomaly detection tasks. Currently, many scholars have used adaptive methods such as Peak Over Threshold (POT) \cite{RF125} to dynamically adjust the anomaly threshold and decision boundary. In conclusion, solving this problem requires more accurate model estimation with stronger representation learning ability.

\subsubsection{Challenges of Transformer-based models}

\paragraph{The application of Transformer in OOD tasks} \hfill

From the previous part, the current studies have only made some early attempts on the OOD task, and have not put forward substantially innovative methods/models. Although these studies have demonstrated that Transformer is more effective than other NLP models (i.e., LSTM/Bag-of-words model) through different experimental protocols, they have also indicated that Transformer still has many limitations and deficiencies, which need further research and development.

\paragraph{Multi-modal anomaly detection} \hfill

In anomaly detection tasks, the misclassification of anomalous samples is usually much more costly than the misclassification of normal instances. However, current approaches are mainly targeted at anomaly detection tasks with a single data source, even MTS data tasks are also aimed at detecting multiple data dimensions of the same system. In contrast, many anomalies exist in multi-modal data, such as combining two or more audio, video, graphics, image, and text data simultaneously. The multi-modal task requires models to consider the correlation and complementarity of different feature distributions over the potential space, which is more difficult to implement, so designing multi-modal anomaly detection systems with practical applications is one of the challenges in this area.

\paragraph{Transformer model optimization} \hfill

Although many methods have been devoted to optimizing the performance overhead of Transformer since it was proposed, such as sparse attention mechanism, Informer, etc., these methods also have their drawbacks, such as restricted application scenarios and no significant time complexity improvement compared with Vanilla Transformer in certain cases. Therefore, how to reduce the resource consumption model is an issue that all Transformer-based anomaly detection models have to consider. Otherwise, these models will be difficult to deploy to edge computing platforms with constrained computability. For example, Ullah et al. \cite{ULLAH2023103289} employed filter pruning via geometric median (FPGM) to compress and reduce the size of the original anomaly detection model. In addition, although Transformer has many variants, there is no universal variant model when dealing with the problem of overly long input data with corresponding position encoding \cite{RF126}. Therefore, Transformer itself still needs continuous iterative updates.

\subsection{Future research trends and development directions}

We believe that the current development directions of Transformer are unsupervised binary anomaly detection with higher performance and model interpretability, and multi-class classification anomaly detection tasks based on semi-supervised and weak-supervised methods. Deeper theoretical research is worth studying. Therefore, researchers should focus on interpretive learning and data-driven sustainability to meet the higher needs of anomaly detection tasks.

\subsubsection{Zero-shot learning}
The meaning of Zero-Shot Learning \cite{RF127} (ZSL) is to enable models to classify categories they have never seen before, giving machines the ability to reason and achieve true intelligence. Therefore, the test set for ZSL can be divided into two categories: the first in which all the test sets are new categories; The second in which the test set includes both existing categories in the training set and new categories. Improperly trained models will conservatively tend to classify new categories as existing categories. To some extent, the binary classification anomaly detection task can be considered as a special case of ZSL. There are many benefits of using ZSL, such as the hierarchical representation structure of the ZSL method helps to eliminate the deviation caused by data imbalance. However, the application of ZSL must address the following issues:

\begin{enumerate}
    \item The effect of ZSL depends on information about similar modalities;
    \item ZSL still lacks an adequate amount of specialized definitions and descriptions.
\end{enumerate}

Therefore, there are few studies combining Transformer with ZSL methods. Wang et al. \cite{RF35} have tested the performance of Transformer in ZSL tasks and Few-Shot Learning (FSL) tasks. Pillai et al. \cite{9897615} investigated the use of Transformer in the task of video anomaly detection under few-shot learning conditions, considering the first few consecutive non-anomalous frames of the video. They utilized features extracted from the frames to train the Transformer in predicting the non-anomalous features of the subsequent frame. Consequently, this paper predicts that in the future, Transformer will increasingly leverage ZSL method to enhance its representation and model generalization capabilities.

\subsubsection{Explainable learning}

The target of explainable learning is to make the training process and experimental results of neural networks logically explainable. This problem not only exists in Transformer but also in GAN, GNN, and other neural networks. For example, hyperparameter selection, nonlinear activation function selection, and threshold selection for anomaly detection in neural networks should be explainable to support human experts’ decision-making and reasoning. Zhang et al. \cite{RF108} used an explainable method in their paper to select the optimal threshold for anomaly detection. Marino et al. \cite{RF60} showed their desire to explore existing explainable algorithms (such as LIME and SHAP) to provide explanations at the packet level and byte level. Explainability enables us to understand the meaning of the representations that the model learns from the samples and to control the learning process more purposefully. Explainability also detects and eliminates edge cases. However, current researchers do not pay much attention to the explainability of Transformer, so it is one of the key issues in this field to propose more interpretable models for anomaly detection.

\subsubsection{Life-long learning}

In some anomaly detection domains, life-long learning is necessary, such as anomaly detection in the ADS field, because the traffic information on urban roads is always in a real-time change and never-ending state, which requires models with life-long learning capabilities to continuously improve the reliability of anomaly detection. In addition, life-long learning also has the following advantages: it enables the model to retain the detection capability for the original task while adjusting and expanding the model for more complex tasks, improving old knowledge, and storing new knowledge for future use. Han et al. \cite{RF80} introduced life-long learning in their proposed ADS anomaly detection system to continuously collect data and improve the model in real time. Li et al. \cite{RF59} aimed to utilize lifelong learning to tackle the issue of false alarms generated by fine-tuned models when encountering forgotten normal patterns. In the future, the development trend of life-long learning should aim to eliminate the need for retraining the model with additional training samples. Instead, the focus should be on incremental learning, where the representation learning of new samples is considered. Ensuring effective life-long learning in the presence of unknown datasets becomes one of the crucial research directions in this field \cite{RF128}.

\subsubsection{More efficient unsupervised binary classification anomaly detection and semi-supervised / weak-supervised multi-class classification anomaly detection}\label{7.3.4}

At present, most of Transformer's existing research in the field of anomaly detection is supervised multi-class classification anomaly detection and unsupervised binary classification anomaly detection. Relevant studies of various anomaly detection are shown in Figure \ref{figure11}:

\begin{figure}[H]
    \centering
    \includegraphics[width=1.0\linewidth]{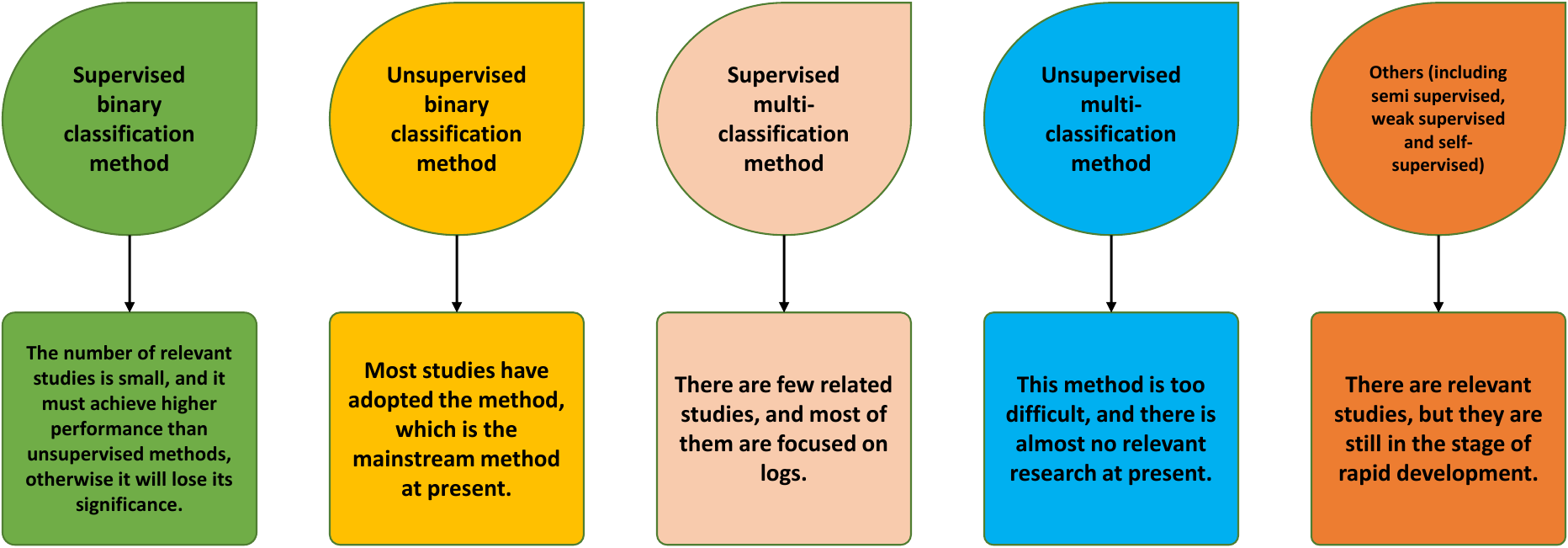}
    \caption{Types and characteristics of various anomaly detection related studies}
    \label{figure11}
\end{figure}

As can be seen from Figure \ref{figure11}, there are basically no unsupervised multi-class classification anomaly detection methods. In this paper, we believe that researchers should strive to improve the performance of unsupervised binary classification anomaly detection tasks, such as enhancing the training efficiency of Transformer, improving input representation, enhancing feature extraction ability, etc. On the other hand, supervised multi-class classification anomaly detection methods should be gradually transformed into semi-supervised and weak-supervised methods. Since the phenomenon of lacking-labeled data is still prevalent, using semi-supervised or weak-supervised learning can further enhance the universality and generalization ability of the model. At the same time, multi-class classification anomaly detection models should be combined with the MAML method, so that they can still effectively learn the features of various types of anomalies even on labeled datasets with severely limited data volume. Using the data enhancement method to expand the multi-class classification dataset is also another feasible approach. Researchers can learn from the idea of manually generating corrupted image samples in the field of image anomaly detection, and artificially generating different types of anomaly samples to expand the multi-class classification dataset and enhance the stability of model training. Therefore, using the MAML method, manual data expansion, and using semi-supervised \& weak-supervised training methods are several important directions for the development of multi-class classification anomaly detection models.

\subsubsection{Method fusion and hybrid model}
It is not difficult to see from the previous summary of scientific research results that the fusion of methods and the use of hybrid models are almost an inevitable trend in the development of anomaly detection. No neural network model can provide perfect, unassailable performance and efficiency for anomaly detection tasks, and so does Transformer. Hybrid models enable Transformer to complement other models for higher performance.

\subsubsection{Contrastive Learning}

Contrastive learning is essentially a self-supervised learning method, which lets models learn the general features by learning the similarities or differences of data points. Contrastive learning can enhance the feature difference between normal and anomalous samples, which is of great significance for anomaly detection in the context of large unlabeled datasets. In addition, contrastive learning can reduce the knowledge cost required for the model to discriminate anomalies by identifying them with only some key features in the hidden space. However, the applications of contrastive learning in the current Transformer-based anomaly detection model are very few. Wang \cite{RF33}, Fan et al. \cite{RF77} and Zhou et al. \cite{RF96} introduced contrastive learning in their models. Therefore, for Transformer anomaly detection models using self-supervised methods, the use of contrastive learning is one of the important future research directions.

\subsubsection{New application scenarios}

In Section \ref{5}, this paper introduces the application scenarios of the Transformer-based anomaly detection model. Since there is a consensus in the artificial intelligence community that Transformer is a generalized model, we believe that Transformer should be applied in more anomaly detection scenarios in the future, rather than the fields of image, audio, sequence data, etc. This is also supported by the other categories in the application scenario in Section \ref{5}. In the future, we hope that Transformer model can be applied to more global real-time hotspot tasks, such as anomaly identification and trend prediction of pandemic virus.

\section{Conclusion} \label{8}

In this paper, we present a comprehensive review of anomaly detection methods based on the Transformer architecture. We delve into the theoretical and methodological aspects of utilizing the Transformer in anomaly detection tasks. Moreover, we summarize and analyze a range of methods, considering their applicability to different scenarios and evaluating their performance. Additionally, the paper identifies key research issues, challenges, and outlines future directions for further exploration in this area. We hope that this review will help readers gain a comprehensive understanding of Transformer's research in the field of anomaly detection. Since the research in this area is still in its infancy, we are confident that more innovative techniques and frontier theories based on Transformer will emerge in the future.

\section*{Acknowledgement}

The authors thank the anonymous reviewers for their insightful suggestions on this work.

\section*{Declarations}

\begin{itemize}
\item Funding: This work was supported by the National Key Research and Development Program of China Under Grant Nos (2022YFB3103403), the Fundamental Research Funds for the Central Universities Under Grant Nos (YCJJ20230464), and the National Natural Science Foundation of China Under Grant Nos (6217071437, 62072200, 62127808).
\item Competing interests: The authors have no competing interests to declare that are relevant to the content of this article.
\item Availability of data and materials: No datasets were generated during the current study.
\item Authors' contributions: Mingrui Ma: Conceptualization, Methodology, Software, Formal analysis, Data Curation, Writing - Original Draft, Validation, Writing - Review \& Editing; Lansheng Han: Investigation, Visualization, Resources, Funding acquisition; Chuanjie Zhou: Supervision, Project administration.
\end{itemize}

%% \noindent
%% If any of the sections are not relevant to your manuscript, please include the heading and write `Not %% applicable' for that section. 



\bibliography{reference}% common bib file

\end{document}